\documentclass{article}
\usepackage{amsmath,epsfig}
\usepackage[preprint]{spconfa4}
\usepackage{xcolor}
\usepackage{cite}

\usepackage{amsfonts}
\usepackage{mathrsfs}
\usepackage{bm}
\usepackage[misc]{ifsym} 

\usepackage{amssymb}
\usepackage{subfigure}
\graphicspath{{figures/}}
\usepackage{multirow} 

\let\OLDthebibliography\thebibliography
\renewcommand\thebibliography[1]{
  \OLDthebibliography{#1}
  \setlength{\parskip}{0pt}
  \setlength{\itemsep}{0pt plus 0.3ex}
}

\begin{document}\sloppy

\def\x{{\mathbf x}}
\def\L{{\cal L}}

\title{Attention Distraction: Watermark Removal Through Continual Learning with Selective Forgetting}
%
\name{Qi Zhong$^{\ast}$, Leo Yu Zhang$^{\ast}$$^{(\textrm{\Letter})}$, Shengshan Hu$^{\dagger}$, Longxiang Gao$^{\ddagger}$$^{\S}$, Jun Zhang$^{\#}$, Yong Xiang$^{\ast}$}
\address{$^{\ast}$Deakin University;
$^{\dagger}$Huazhong University of Science and Technology; 
$^{\ddagger}$Qilu University of Technology;\\
$^{\S}$Shandong Computer Science Center; 
$^{\#}$Swinburne University of Technology \\ email: leo.zhang@deakin.edu.au}


\maketitle

\begin{abstract}  
Fine-tuning attacks are effective in removing the embedded watermarks in deep learning models.
However, when the source data is unavailable, it is challenging to just erase the watermark without jeopardizing the model performance. 
In this context, we introduce Attention Distraction (AD), a novel source data-free watermark removal attack, to make the model selectively forget the embedded watermarks by customizing continual learning.
In particular, AD first anchors the model's attention on the main task using some unlabeled data.
Then, through continual learning, a small number of \textit{lures} (randomly selected natural images) that are assigned a new label distract the model's attention away from the watermarks.
Experimental results from different datasets and networks corroborate that AD can thoroughly remove the watermark with a small resource budget without compromising the model's performance on the main task, which outperforms the state-of-the-art works.
\end{abstract}
\section{Introduction}
\label{sec:intro}
Deep neural network (DNN) watermarking, deriving wisdom from digital watermarking, protects the intellectual property of DNN models by endowing the model with the ability to trace illegal distributions. Since the first work proposed by Uchida \textit{et al.} \cite{uchida2017embedding} in 2017, researchers have given a lot of enthusiasm to this field and a variety of solutions have been put forward \cite{adi2018turning, zhang2018protecting, zhong2020protecting, li2020protecting}.
The vulnerabilities of watermarks in DNN, meanwhile, have also attracted the attention of researchers. 

The watermark removal attack based on fine-tuning technology is one of the most serious threats to watermarked models.
It aims to make the victim model forget the embedded watermarks but preserve accuracy on the main task through fine-tuning the model with source training data or/and proxy data (unrelated data acts as a substitution or augmentation of the source training data) \cite{chen2021refit, liu2020removing, aiken2021neural, wang2021detect}.
However, it is nontrivial to achieve such a problem containing two conflicting goals in the source data-free regime due to the catastrophic forgetting phenomenon \cite{kemker2018measuring}  of deep learning systems.

The work in \cite{shafieinejad2019robustness} is the first to explore the feasibility of watermark removal in DNN using unlabeled data obtained from open sources. 
Its improved version, proposed in \cite{guo2021hidden}, uses a dataset transformation method called PST (Pattern embedding and Spatial-level Transformation) to preprocess the data before model fine-tuning.
However, they either have no effect on particular watermarks or impair the model's performance after erasing the watermark.

\begin{figure}
\centering
  \includegraphics[scale=0.38]{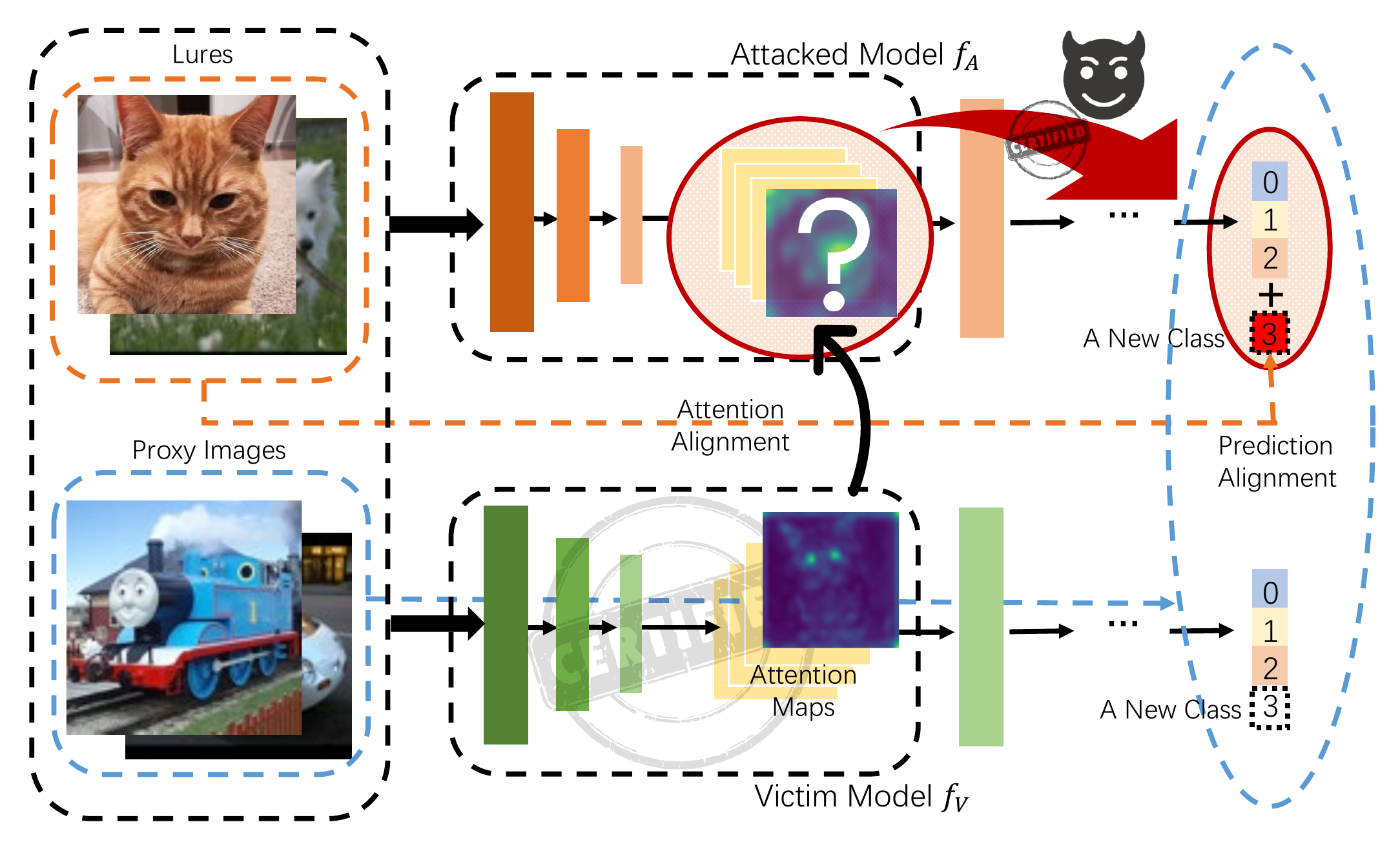}
  \caption{The framework of AD. It first assigns a new class to the lure samples and feeds them into the victim model, which motivates the model to continue learning the lure task by minimizing the cross-entropy loss. It then fine-tunes unlabeled proxy samples by minimizing attention alignment and prediction alignment losses.}
  \label{fig:framework}
 \end{figure}
To mitigate this gap and solve the selective forgetting watermarks (\textbf{SFW}) problem in DNNs, we propose a novel continual learning-based  \cite{li2017learning} DNN watermark removal attack in the source data-free regime: Attention Distraction (\textbf{AD}). As depicted in Fig.~\ref{fig:framework},
AD takes a small part of out-of-distribution (OOD) unlabeled samples as \textit{lures} and assigns them an additional new class to distract the victim's attention from the watermark task. The lures can be easily collected from open sources.
Meanwhile, in the fine-tuning process, AD anchors model attention on the main task by aligning the intermediate-layer attention maps and the final softmax outputs of proxy data with that of the original victim model.
The rationale behind this is simple but effective: 
through continual learning, distracting the model's attention from watermarks endows the model with the ability to selectively forget the knowledge about watermarks.

In summary, we make the following contributions:
\begin{itemize}
    \item We propose AD, a novel continual learning-based DNN watermark removal framework that leverages attention anchoring and attention distraction strategies to achieve SFW, and present intuitive explanations for its efficacy. 
    \item 
    Unlike previous works that require access to source training data, computationally intensive, or only applicable to specific watermark types, AD requires no source training data and is resource-efficient as well as agnostic to watermarking mechanism types.
    \item Comprehensive empirical experiments on various benchmark datasets and network architectures validate that AD can effectively remove existing watermarks. 
\end{itemize}

\section{Background and Related Works}
\subsection{DNN Watermarking}
\label{subsec: dnn_wm}


This work focuses on the query-response-based watermarking \cite{adi2018turning, zhang2018protecting, zhong2020protecting, li2020protecting}, which is built on a more realistic assumption that only API access to the to-be-protected model is required.  Mathematically, the general process of query-response-based DNN watermarking can be characterized as follows.

Considering a victim $\mathcal{V}$ who trains a classifier $f_{\mathcal{V}}$ on a clean training data $\mathcal{D}=\{\bm{x}^{(i)}, y^{(i)}\}_{i=1}^{N}$ $(y^{(i)}\in \{0,1,...,C-1\})$.
The watermark is embedded into the model by minimizing the following loss function:
\begin{equation}
\mathcal{L}(\bm{\theta}_\mathcal{V}) = \mathbb{E}_{{\bm{x}\in \mathcal{D}\cup \mathcal{T}}} \left[\mathcal{L}_{CE}(f_{\mathcal{V}}(\bm{x}; \bm{\theta}_\mathcal{V}) , y)\right] + \alpha \cdot \mathcal{R}(\bm{\theta}_\mathcal{V}), 
\label{eq:ce_loss}
\end{equation} 
where $\mathcal{T}=\{\bm{x}^{(i)}_{t}, y_t\}_{i=1}^{M}$ is a self-designed trigger set, $y_t \in \{0,1,...,C-1\}$ is the target label selected from existing classes, $\bm{\theta}_\mathcal{V}$ represents all trainable parameters, $\mathcal{L}_{CE}$ denotes the cross-entropy loss, and $\alpha$ is a parameter that controls the strength of the regularization term $\mathcal{R}(\bm{\theta}_\mathcal{V})$ to prevent overfitting.

If $\mathcal{V}$ suspects model theft, to confirm whether the suspected model contains the watermark, she queries this API and checks the accuracy 
\begin{equation}
Acc=\mathsf{Ver}_{_{\bm{x}\in \mathcal{T}}}(f_{\mathcal{V}}(\bm{x}; \bm{\theta}_\mathcal{V}), y_t),
\end{equation}
where $\mathsf{Ver}$ is the verification function returns the probability for $f_{\mathcal{V}}(\bm{x}; \bm{\theta}_\mathcal{V})= y_t$.
If $Acc>\tau$ ($\tau$ is a threshold close to 1), $\mathcal{V}$ can claim her ownership of the model.

\subsection{Watermark Removal Attacks in DNN}
\label{subsec: wm_removal}

Previous researchers explored various fine-tuning-based attacks to remove DNN watermarks, which can be roughly divided into three main types according to the resources budget they can apply: the source data available regime \cite{adi2018turning, zhang2018protecting}, the source data limited regime \cite{chen2021refit, liu2020removing, aiken2021neural, wang2021detect}, and the source data-free regime \cite{shafieinejad2019robustness, guo2021hidden}.
The following briefly introduces the fine-tuning based watermark removal attacks targeted at the source data-free regime.

The authors in \cite{shafieinejad2019robustness} proposed a vanilla attack that employs unlabeled data obtained from open source as proxy data and annotates them using the original victim model as a labeling oracle. However, it causes the model's performance on the main task to degrade since the pseudo labels of the proxy data provided by the model contain noise.
To mitigate this drawback, Guo \textit{et al.} in \cite{guo2021hidden} proposed PST-FT attack, which combines an elaborately designed data transformation method called PST with fine-tuning. 
It is mainly based on the assumption that the fragile mapping between the triggers and the target label learned by the victim model can be easily destroyed after fine-tuning using the PST-transformed dataset.
In the ownership verification phase, the clean instances processed by PST can still be recognized by the attacked model, while the triggers after the PST processing become unrecognizable.

\section{Problem Statement}


\subsection{Attack Setting}

The attacker $\mathcal{A}$ has obtained the victim model $f_{\mathcal{V}}$, he is aware of the existence of watermarks, but he neither knows the watermarking method nor the knowledge of triggers, \textit{e.g.,} the type of the triggers or the target label. He has a certain power of computing resources, but he cannot access the source training data $\mathcal{D}$. In other words, the attacker has no knowledge of the watermarks or the source training data, but he can access the victim model and do some manipulations on it.





\subsection{Problem Formulation}

We consider a victim model $f_{\mathcal{V}}$, which is watermarked by optimizing the object Eq.~(\ref{eq:ce_loss}) using the clean dataset $\mathcal{D}$ and the trigger set $\mathcal{T}$. 
We assume the victim model $f_{\mathcal{V}}$ has the state-of-the-art performance on both the main task (\textit{i.e.,} classification on the clean dataset $\mathcal{D}$ ) and the watermark task (classification on the trigger set $\mathcal{T}$).
The constraints are that only a certain number of auxiliary data $\mathcal{D}_A$ are available in the AD attack, while the triggers or watermarking methods are agnostic. 

The attacker aims to eliminate the watermark from the victim model.
%
We view SFW as an instance of multi-task learning with conflicting objectives. Namely, using a continual learning approach with a limited resource budget to find an optimal self-balancing loss function that achieves high accuracy on the main task but low accuracy on the watermark task.
In summary, to get the optimum classifier $f_{\mathcal{A}^*}$, the attack can optimize
\begin{eqnarray} 
\mathop{\text{min}}\limits_{\bm{\theta}_{\mathcal{A}}} ~\mathcal{L}_{AD}(\bm{\theta}_{\mathcal{A}}),~\text{for}~\bm{x}\in \mathcal{D}_{A},  
\end{eqnarray}
such that the following equations are hold:
\begin{eqnarray} 
|\text{Prob}_{\bm{x}\in \mathcal{T}}[f_{\mathcal{A}^*}(\bm{x})= y_t]-\frac{1}{C}| \leq \epsilon \label{eq:wm_forget},\\
|\text{Prob}_{\bm{x}\in \mathcal{D}}[f_{\mathcal{A}^*}(\bm{x})= y] - 
\text{Prob}_{\bm{x}\in \mathcal{D}}[f_{\mathcal{V}}(\bm{x})= y]| \leq \text{negl},\label{eq:fidility} 
\end{eqnarray}
where $\mathcal{L}_{AD}$ is the desired objective loss function for the attacker, $\text{negl}$ is a negligible value and $\epsilon$ determines to what extent the fine-tuned model forgets the original watermark.

\section{Attention Distraction Attack}
\label{sec:ad_attack}

\subsection{The Design Rationale}
\label{sec:rationale}



Recent studies on DNN visualization \cite{zeiler2014visualizing, selvaraju2017grad} find that activation maps can be used to build heatmaps (\textit{i.e.}, attention maps) that imply models' attention on images. And different models often share similar attention when making the right decisions for the same inputs \cite{wu2020boosting}.
Intuitively, in the fine-tuning/continual learning process, if we can maintain the attention maps for the main task while disrupting the attention maps for the watermark task, then we can achieve SFW.
\begin{figure}[!t]
\centering
  \includegraphics[scale=0.2]{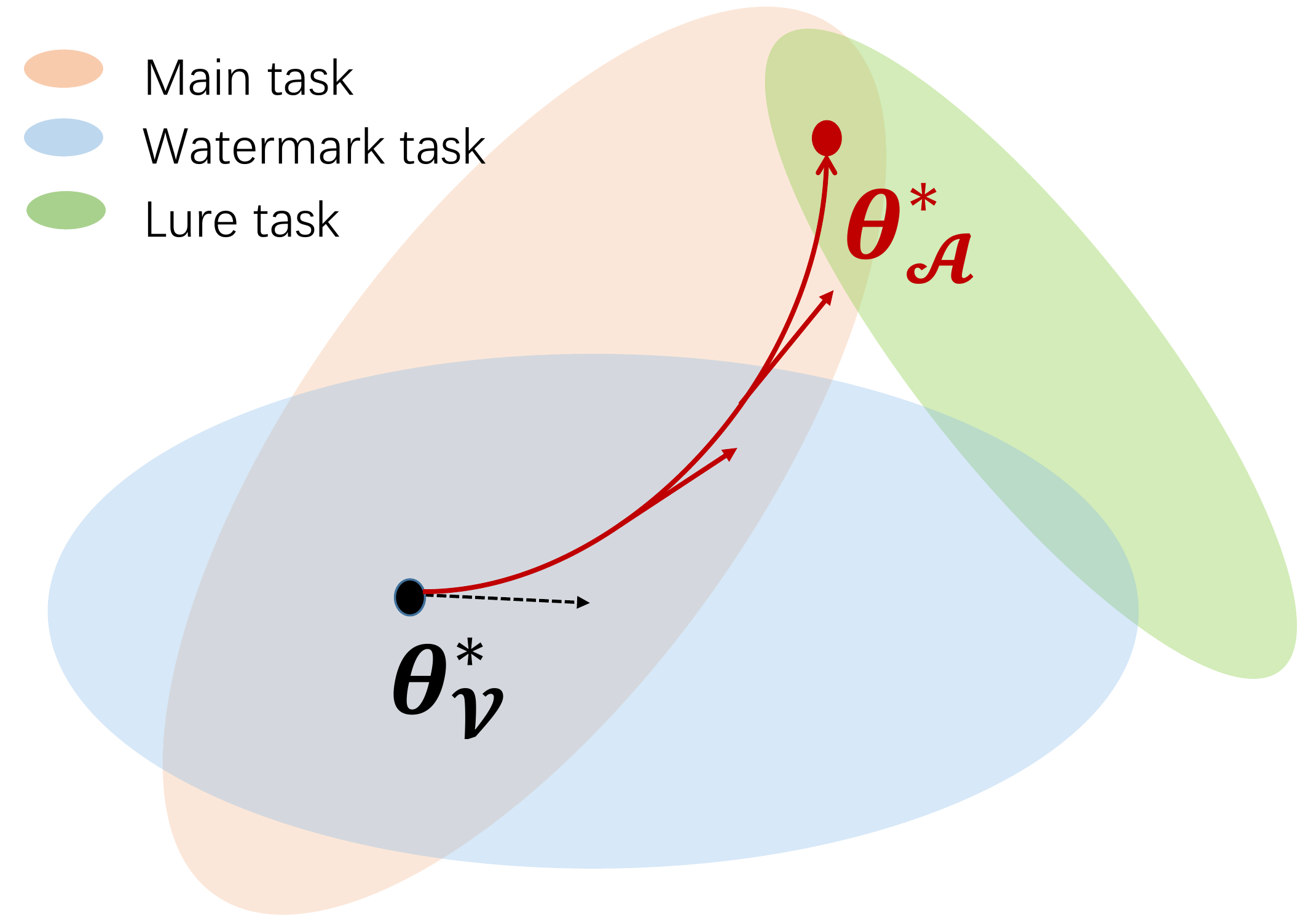}
  \caption{Conceptual diagram for selective forgetting of the watermark task. Attention distraction happens whilst learning the lure task. The overlapping region of multiple ellipses represents the common solution space for those tasks.}
  \label{fig:att_dis}
 \end{figure}

As shown in Fig.~\ref{fig:att_dis}, initially, the optimum $\theta_{\mathcal{V}}^{*}$ comes with acceptable error allowed by both the main task and original watermark task.
The red arrows direct the path to update weights whilst learning the lure task. 
The continual learning of the lure task prompts the model weights to update in the direction that is conducive for the classification of the lures.
The design of the lure task should ensure that it does not overlap with the watermark task, so it is more likely for the model to find a path that is only conducive for the classification of the main task and the new task, \textit{i.e.}, the red arrows' direction. 
At the end, it can reach an optimum $\theta_{\mathcal{A}}^{*}$ with acceptable error fits for both the main task and lure task but not the original watermark task.

\subsection{Auxiliary Dataset Crafting}
As we have mentioned above, for scarce of source training data, to selectively forget watermarks, we need assistance from a certain number of auxiliary data, \textit{i.e.}, proxy samples and lure samples.  

In our method, the unlabeled proxy samples $\mathcal{D}_{P}=\{\bm{x}^{(i)}_{p}, \perp\}_{i=1}^{N_{P}}$ can be collected from the Internet, drawn from the in-distribution (ID) or OOD area of the source training data. 
The lure dataset $\mathcal{D}_{L} $ can be drawn from the OOD area of the source training data, \textit{e.g.}, randomly selected from natural images. 
To prevent feature collision between proxy data and lure data, there should be no intersection between them (\textit{i.e.}, $\mathcal{D}_{L} \cap \mathcal{D}_{P} = \emptyset$) and their distributions are different.
Then a new label $\Delta~(\Delta = C)$ is assigned to the lure samples, and the whole lure set is $\mathcal{D}_{L}=\{\bm{x}^{(i)}_{l}, \Delta\}_{i=1}^{N_{L}}$. And the proxy samples and lure samples consist of the required auxiliary data, \textit{i.e.},  $\mathcal{D}_{A}= \mathcal{D}_{P} \cup \mathcal{D}_{L}$.

\subsection{Watermark Removal}

Since the lure data $ \mathcal{D}_{L}$ is a new task for the victim model, manipulations of the model are necessary before the continual learning process.
We add a new class to the output layer of $f_{\mathcal{V}}$ and make it fully connected to the layer beneath. 
The ground-truth label of the lures is $\Delta=C$, we formulate the prediction loss function for the lures as
\begin{eqnarray} 
\mathcal{L}_\delta = \mathbb{E}_{_{\bm{x}\in \mathcal{D}_{L}}}\left[\mathcal{L}_{CE}(f_{\mathcal{A}}(\bm{x}; \bm{\theta}_{\mathcal{A}}),~ C)\right].
\label{eq:pred_lu}
\end{eqnarray}

The authors in \cite{adi2018turning} have demonstrated that fine-tuning or retraining the deep layers (\textit{i.e.}, the fully-connected layers) only has little effect to remove the watermark. Based on this consideration, we fine-tune the whole network. We rewrite the victim model into a composite function
\begin{eqnarray}
    f_{\mathcal{V}}(\bm{x}; \bm{\theta}_{\mathcal{V}}) = \mathsf{F}^{(2)}( \mathsf{F}^{(1)}(\bm{x};\bm{\theta}^{(1)}_{\mathcal{V}}); \bm{\theta}^{(2)}_{\mathcal{V}}),
\end{eqnarray}
in which 
$\mathsf{F}^{(1)}:\mathbb{R}^{H\times W \times C}\rightarrow  \mathbb{R}^{H_{l}\times W_{l} \times C_{l}}$ denotes the attention mapping function that propagates an input $\bm{x}$ through the network to the $l$-th ($l \in [1,2,\cdots, L]$) activation layer,
$\mathsf{F}^{(2)}:\mathbb{R}^{H_{l}\times W_{l} \times C_{l}}\rightarrow  \mathbb{R}^{C+1}$ propagates the attention maps from $\mathsf{F}^{(1)}$ to the softmax output layer.
In our approach, we select a relatively deep layer, \textit{e.g.}, the penultimate activation layer, to construct the sub-net $ \mathsf{F}^{(1)}$.

Since the source victim model $f_{\mathcal{V}}$ contains watermarks, their attention maps on the proxy data will inevitably, more or less, carry the watermark information.
Accordingly, we penalize the attention maps of the model on proxy data and formulate our attention anchoring loss function as 
\begin{equation}
\mathcal{L}_{AA} = \lambda_{1}\cdot\mathcal{L}_{v} +\lambda_{2}\cdot\mathcal{L}_{a} + \lambda_{3}\cdot\mathcal{R}(\bm{\theta}_{\mathcal{A}}),
\label{eq:loss_aa}
\end{equation} 
where $\lambda_{i}$ is the coefficient that regulates the importance of the item,
$\mathcal{L}_{v}$ is the vanilla prediction loss term defined as
\begin{equation}
\mathcal{L}_{v} = \mathbb{E}_{{\bm{x}\in \mathcal{D}_{P}}}\left[\mathcal{L}_{KL}(f_{\mathcal{A}}(\bm{x}; \bm{\theta}_{\mathcal{A}}),~ f_{\mathcal{V}}(\bm{x}; \bm{\theta}_{\mathcal{V}}))\right],
\label{eq:loss_v}
\end{equation} 
$\mathcal{L}_{KL}$ is the Kullback-Leibler divergence,
$\mathcal{L}_{a}$ is the attention alignment loss term
\begin{eqnarray} 
\mathcal{L}_{a} = \mathbb{E}_{_{{\bm{x}\in \mathcal{D}_{P}}} } \left[\Vert \mathsf{F}^{(1)}(\bm{x};\bm{\theta}_{ \mathcal{A}}) -\mathsf{F}^{(1)}(\bm{x};   \bm{\theta}_{\mathcal{V}})\Vert_{2}\right],
\label{eq:L_a}
\end{eqnarray}
and $ \mathcal{R}(\bm{\theta}_{\mathcal{A}})$ is the attention penalty term for the proxy data
\begin{eqnarray} 
 \mathcal{R}(\bm{\theta}_{\mathcal{A}})=\mathbb{E}_{{{\bm{x}\in \mathcal{D}_{P}}}}\left[ \Vert\mathsf{F}^{(j)}(\bm{x}; \bm{\theta}_{\mathcal{A}}) \Vert_{2}\right],\
 \label{eq:R_a}
\end{eqnarray}
for $j = 1, 2$. To summarize, the total loss for solving the SFW problem with AD is
\begin{eqnarray} 
\mathcal{L}_{AD} =
\mathcal{L}_{\delta} + \mathcal{L}_{AA}.
\label{eq:finalloss}
\end{eqnarray}

\section{Experiments}

\subsection{Experiment Setting}
All the experiments are performed on a PC equipped with a NVIDIA Geforce RTX 2070 GPU and Ubuntu 20.04 OS, and Keras is the underlying framework.

\begin{figure}[htbp]
\centering
\subfigure[]{
\includegraphics[width=1.9cm]{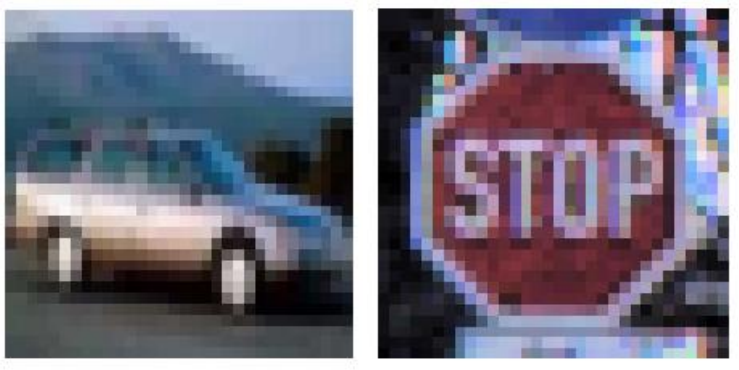}
}
\subfigure[]{
\includegraphics[width=1.9cm]{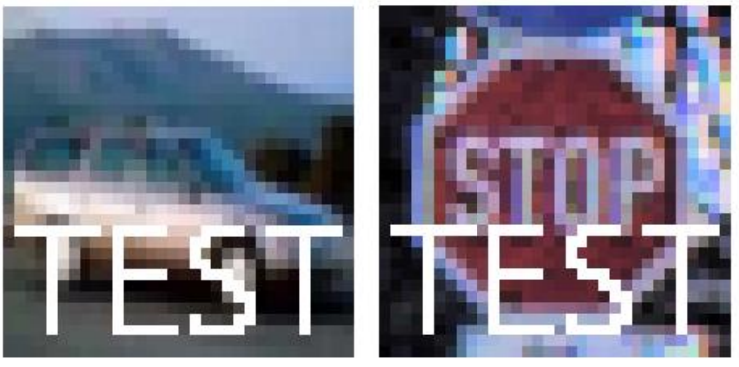}
}
\subfigure[]{
\includegraphics[width=1.9cm]{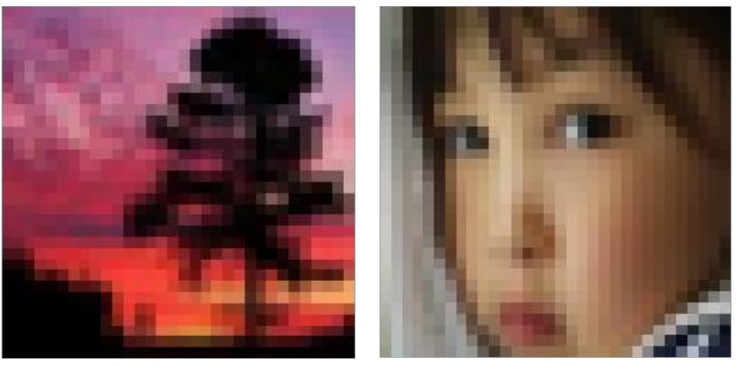}
}
\subfigure[]{
\includegraphics[width=1.9cm]{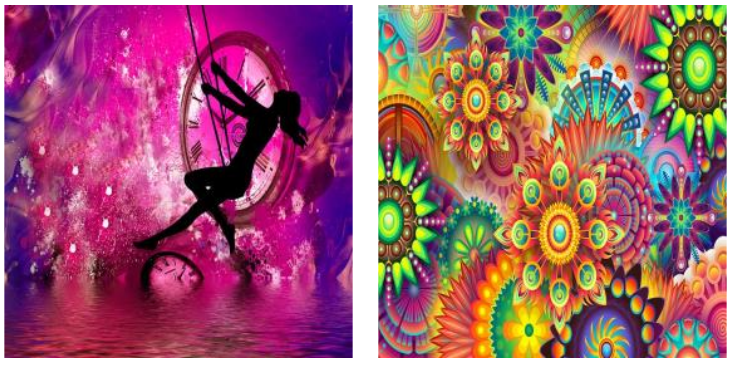}
}
\caption{Examples of datasets: (a) source training samples from CIFAR-10 and GTSRB, (b) trigger samples using `TEST' as perturbation, (c) proxy instances drawn from the CIFAR-100 dataset, and (d) lure images drawn from abstract images.}
\label{fig:samples}
\end{figure}

In our experiments, the watermarked models are mainly targeted at classifying CIFAR-10  \cite{krizhevsky2009learning} and GTSRB \cite{stallkamp2012man}. 
The implemented network architectures of victim models include VGG16 \cite{simonyan2015very}, ResNet18 \cite{he2016deep}, and WRN-16-4 (Wide Residual Network) \cite{zagoruyko2016wide}.
We evaluate the attack effectiveness of AD on three state-of-the-art watermarks introduced in \cite{zhang2018protecting}, 
including the content watermark (as shown in Fig.~\ref{fig:samples}(b)), the noise watermark, and the unrelated watermark.
We set the number of triggers used to embed watermarks into victim models as $1000$, and they are assigned to class $0$ (the target label). 

We consider two attack scenarios for obtaining the proxy datasets: using OOD data (\textit{i.e.}, CIFAR-100 \cite{krizhevsky2009learning}) and ID data (half of the original testing data).
For each case of the obtained proxy datasets, $1000$ images are randomly sampled and used in the fine-tuning. 
In addition, only $10$ randomly select abstract images (introduced in \cite{adi2018turning}) are used as the lure data (the implementation details can be found in the appendix).  

\noindent \textbf{Evaluation metrics.}
We mainly consider the main task accuracy (\textbf{MTA}) and watermark accuracy (\textbf{WMA}) metrics for evaluation. It is performed for the top-1 predicted categories.
For CIFAR-10, we set $\epsilon$ in Eq.~(\ref{eq:wm_forget}) to $0.1$ since $C=10$.

\subsection{Results Analyses}
\label{sec:attack_effect}

\begin{table*}[]
\caption{Performance of watermark removal attacks on the CIFAR-10, where x/y denotes MTA/WMA.}
\small
\label{tab:c10_results}
\centering
\begin{tabular}{|c|c|c|c|cc|cc|}
\hline
  \multirow{2}{*}{Watermark task} &
  \multirow{2}{*}{Networks} &
  \multirow{2}{*}{Victim models} &
  \multirow{2}{*}{PST} &
  \multicolumn{2}{c|}{PST-FT} &
  \multicolumn{2}{c|}{AD} \\ \cline{5-8} 
   &
   &
   &
   &
  \multicolumn{1}{c|}{OOD} &
  ID &
  \multicolumn{1}{c|}{OOD} &
  ID \\ \hline
\multirow{3}{*}{Content} &
  VGG16 &
  93.82 / 99.96 &
  88.20 / 14.84 &
  \multicolumn{1}{c|}{84.70 / 19.14} &
  82.80 / 17.80 &
  \multicolumn{1}{c|}{\textbf{93.58} / \textbf{7.70}} &
  93.38 / 18.28 \\
 &
  ResNet18 &
  93.74 / 99.78 &
  84.60 / 12.44 &
  \multicolumn{1}{c|}{88.66 / 17.98} &
  88.46 / 15.14 &
  \multicolumn{1}{c|}{\textbf{92.76} / 13.94} &
  92.34 / \textbf{6.78} \\
 &
  WRN-16-4 &
  93.42 / 99.94 &
  88.62 / 26.82 &
  \multicolumn{1}{c|}{85.24 / 18.70} &
  88.56 / 26.68 &
  \multicolumn{1}{c|}{\textbf{92.82} / 13.50} &
  91.16 / \textbf{1.98} \\ \hline
\multirow{3}{*}{Unrelated} &
  VGG16 &
  93.68 / 100 &
  88.50 / 4.76 &
  \multicolumn{1}{c|}{86.13 / 16.56} &
  88.26 / 9.34 &
  \multicolumn{1}{c|}{89.56 / 14.78} &
  \textbf{92.40} / \textbf{0.22} \\
 &
  ResNet18 &
  92.88 / 100 &
  84.34 / 12.74 &
  \multicolumn{1}{c|}{88.00 / 1.64} &
  87.90 / 2.12 &
  \multicolumn{1}{c|}{88.88 / 1.16} &
  \textbf{90.02} / \textbf{0.30} \\
 &
  WRN-16-4 &
  93.80 / 100 &
  88.50 / 6.30 &
  \multicolumn{1}{c|}{89.60 / \textbf{1.76}} &
  89.18 / 1.90 &
  \multicolumn{1}{c|}{90.86 / 3.98} &
  \textbf{91.74} / 2.38 \\ \hline
\multirow{3}{*}{Noise} &
  VGG16 &
  93.58 / 100 &
  88.36 / 99.86 &
  \multicolumn{1}{c|}{72.16 / 2.22} &
  73.82 / \textbf{0.34} &
  \multicolumn{1}{c|}{93.78 / 0.78} &
  \textbf{93.94} / 1.32 \\
 &
  ResNet18 &
  93.72 / 100 &
  84.84 / 1.88 &
  \multicolumn{1}{c|}{86.38 / 13.76} &
  85.52 / 18.46 &
  \multicolumn{1}{c|}{93.22 / 5.32} &
  \textbf{93.64} / \textbf{0.96} \\
 &
  WRN-16-4 &
  93.32 / 100 &
  88.58 / 11.92 &
  \multicolumn{1}{c|}{88.22 / 18.32} &
  90.00 / \textbf{2.84} &
  \multicolumn{1}{c|}{93.28 / 14.30} &
  \textbf{93.88} / 3.48 \\ \hline
\end{tabular}
\end{table*}

Table~\ref{tab:c10_results} presents the performance of victim models before and after being attacked by comparing AD with PST-FT on the CIFAR-10 dataset (the results of GTSRB can be found in the appendix).
%
We can observe that almost all three types of watermarks are vulnerable to the PST method. 
In addition, using PST-FT can generally improve the MTA or decrease the WMA of victim models. However, this is not always the case, especially when the triggers' perturbation is noise. 
This is because the trigger noise could be mixed with the distribution mismatch noise. 

For AD, it is clear that all the WMAs of the attacked victim are below $20$\%, achieving the goal of watermark removal. For all the network architectures, AD retains the MTAs of the victim models.  
Moreover, though AD generally performs better with ID proxy data, the differences between MTAs for OOD and ID proxy data are almost always within $2$\%. This fact validates that AD can avoid the noise caused by the data distribution difference between the proxy and original training data while removing the planted watermark.

\begin{table*}[h]
\caption{Performance of watermark removal attacks of AD without using lure data on CIFAR-10.}
\small
\label{tab:vanilla_aa}
\centering
\begin{tabular}{lllll}

\hline
\multicolumn{1}{|c|}{\multirow{2}{*}{Networks}} & \multicolumn{2}{c|}{Vanilla}                       & \multicolumn{2}{c|}{AA}                            \\ \cline{2-5} 
\multicolumn{1}{|c|}{}                          & \multicolumn{1}{c|}{OOD} & \multicolumn{1}{c|}{ID} & \multicolumn{1}{c|}{OOD} & \multicolumn{1}{c|}{ID} \\ \hline
\multicolumn{1}{|c|}{VGG16}    & \multicolumn{1}{c|}{59.96 / \textbf{6.32} }           & \multicolumn{1}{c|}{73.18 / 19.60}           & \multicolumn{1}{c|}{61.52 / 19.94}            & \multicolumn{1}{c|}{\textbf{76.56} / 19.70 }   \\
\multicolumn{1}{|c|}{ResNet18}  & \multicolumn{1}{c|}{66.78 / 19.52}            & \multicolumn{1}{c|}{76.04 / 18.94}           & \multicolumn{1}{c|}{79.10 / \textbf{10.90 }}           &\multicolumn{1}{c|}{ \textbf{85.94} / 18.26}   \\
\multicolumn{1}{|c|}{ WRN-16-4} & \multicolumn{1}{c|}{75.54 / 19.48}            & \multicolumn{1}{c|}{81.40 / 18.50}           & \multicolumn{1}{c|}{50.50 / 18.96}            & \multicolumn{1}{c|}{\textbf{85.18} / \textbf{13.22}}     \\\hline
\end{tabular}
\end{table*}

\subsection{Ablation Study}
In this section, we perform an ablation study to investigate the role of each component of AD on the CIFAR-10 task and the content-based watermark task (more analysis can be found in the appendix).  

\begin{figure}[h]
  \centering
  \includegraphics[width=1\linewidth]{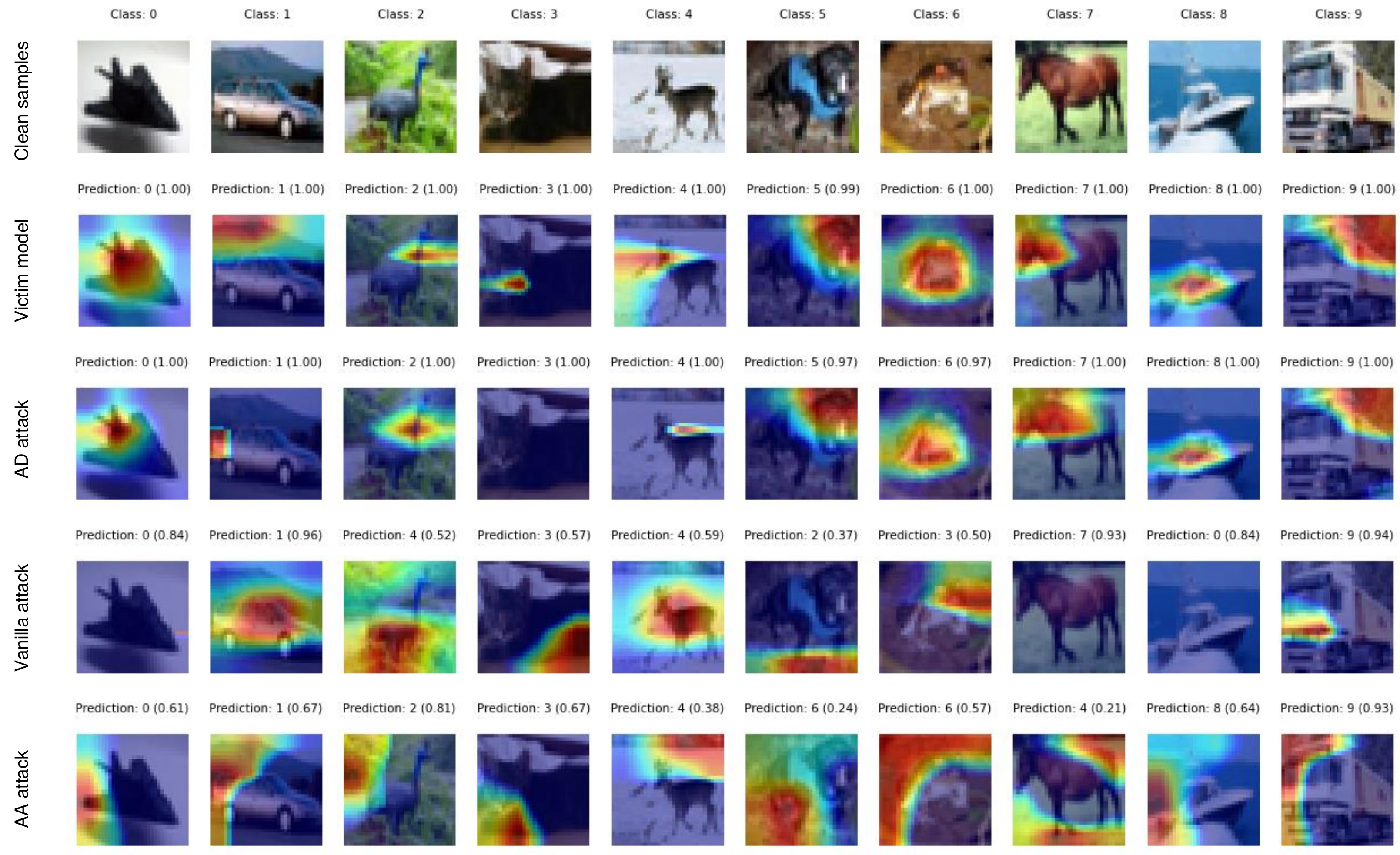}
\caption{The attention heatmaps of ResNet18 on CIFAR-10 samples. 
The first row depicts the clean images from different classes. 
The following rows respectively depict the heatmaps of the victim model before and after the AD attack, vanilla attack, and AA attack with OOD proxy data.}
\label{fig:c_maps}
\end{figure}

\noindent \textbf{Role of the lure task.}
To study the role of the lure task on SFW, we design two experiments that only use proxy data, including OOD and ID, as comparisons.
That said, one method uses the vanilla loss (the vanilla attack proposed in \cite{shafieinejad2019robustness}) $\mathcal{L}_{v}$ (defined in Eq.~(\ref{eq:loss_v})) as the objective, and the other uses attention anchoring (called AA attack) loss $\mathcal{L}_{AA}$ as the objective  (defined in Eq.~(\ref{eq:loss_aa})). And the results are listed in Table~\ref{tab:vanilla_aa}.

From Table~\ref{tab:vanilla_aa}, we observe that, no matter using OOD or ID proxy data, neither the vanilla attack nor the AA attack is as good as AD in solving the SFW.
This confirms that the lure task is important for solving SFW: the lure task will compensate for the negative effect brought by the distribution difference between the proxy data and the original training data. 

To further investigate the role of the lure task, Fig.~\ref{fig:c_maps} visualizes the attention heatmaps (produced by Grad-CAM \cite{selvaraju2017grad}) of ResNet18, before and after different attacks using OOD proxy data, on CIFAR-10 images.
From the second and third rows of this figure, it is easy to see that the attention maps of the model after the AD attack are highly similar to those of the victim before attack. 
For both the vanilla and the AA attacks, the attention maps of the model after attack deviate from those of the victim before attack. 
That said, without the participation of lure, using OOD proxy data alone cannot effectively anchor the main task attention maps while forgetting the target watermark task. 

\begin{table}[h]
\caption{Influence of the lure data label on CIFAR-10.}
\small
\label{tab:lure_tag}
\centering
\begin{tabular}{lllll}
\cline{1-4}
\multicolumn{1}{|c|}{ Networks} & \multicolumn{1}{c|}{$\Delta=0$} & \multicolumn{1}{c|}{$\Delta=1$} & \multicolumn{1}{c|}{$\Delta=5$} &  \\ \cline{1-4}
\multicolumn{1}{|c|}{VGG16} & \multicolumn{1}{c|}{78.70 / 99.34} & \multicolumn{1}{c|}{\textbf{93.58} / 18.78} & \multicolumn{1}{c|}{93.54 / \textbf{6.10} } &  \\
\multicolumn{1}{|c|}{ResNet18} & \multicolumn{1}{c|}{74.88 / 55.90}       & \multicolumn{1}{c|}{92.20 / 18.14}         & \multicolumn{1}{|c|}{\textbf{92.26} / \textbf{11.20}}  \\
\multicolumn{1}{|c|}{WRN-16-4} &\multicolumn{1}{c|}{ 60.70 / 100}         &\multicolumn{1}{c|}{ 93.18 / \textbf{0.00}}          & \multicolumn{1}{|c|}{\textbf{93.76} / \textbf{0.00}}   \\ \cline{1-4}
\end{tabular}
\end{table}

\noindent\textbf{Influence of the label of the lure task.} 
To investigate the influence of the label of the lure task, we conduct experiments using different predefined lure labels, and the results are listed in Table~\ref{tab:lure_tag}.
From this table, we can observe that when the lure label $\Delta$ is not the same as the target label $0$, \textit{i.e.}, $\Delta=1$ or $5$, the results of the AD attack are comparable to those obtained by using a new class $C$ as the label for lure data. 
However, when $\Delta=0$, the AD attack fails for all victim models.

As discussed previously and visualized in Fig.~\ref{fig:att_dis}, when the lure label does not overlap with the target label, it encourages the continual learning process to find a path that is suitable for the main task and the lure task, while forgetting the target watermark task. For the consideration of watermark agnostic, the most convenient way to prevent the label collision is to set $\Delta=C$.

\begin{figure}[h]
  \centering
  \includegraphics[scale=0.4]{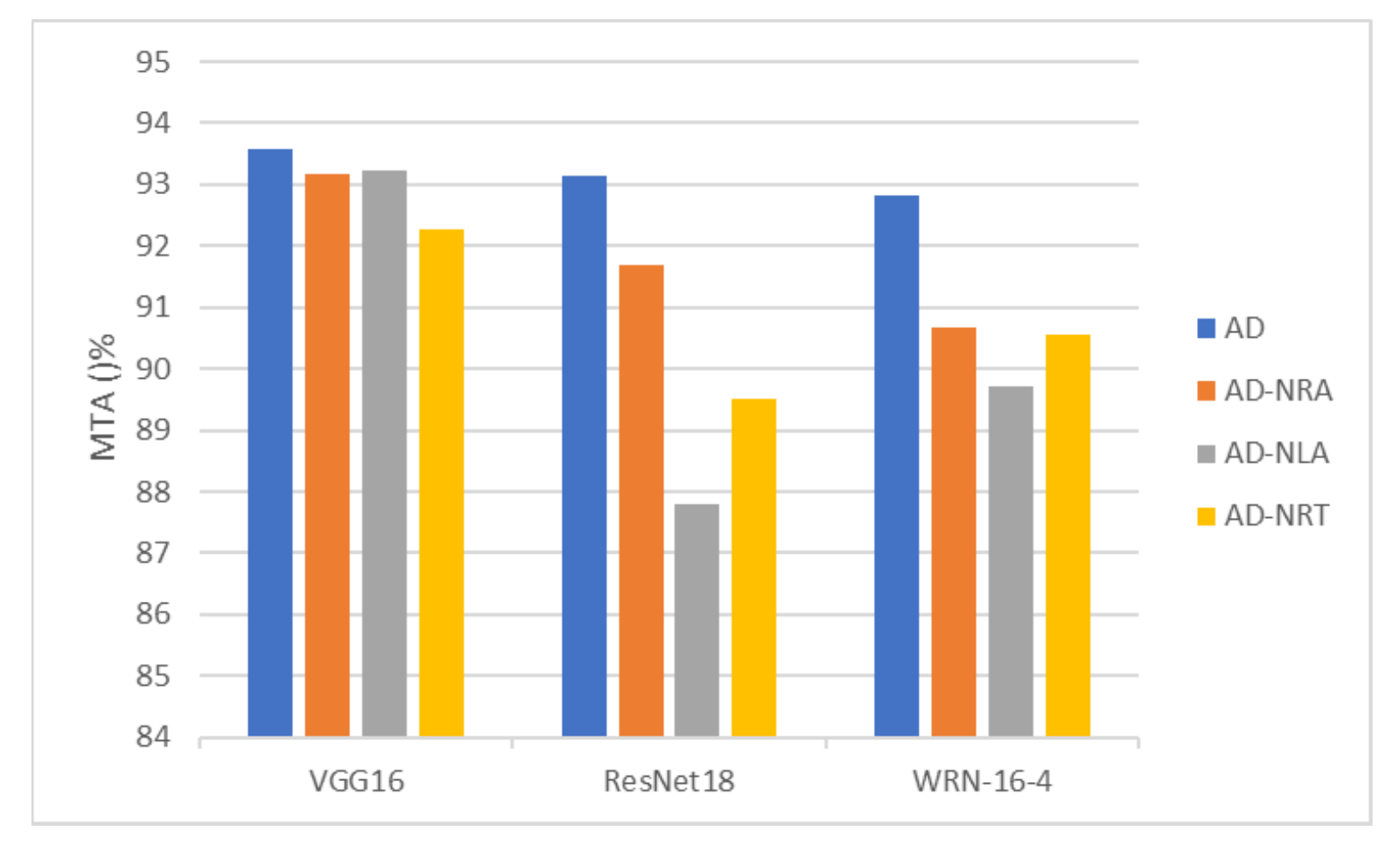}
  \caption{The MTA of AD with different regularization on CIFAR-10. All WMAs are lower than the predefined threshold. The blue bars represent the complete version of AD, the orange bars, gray bars, and yellow bars represent different versions of AD, respectively.}
\label{fig:mat_c10}
\end{figure}

\noindent\textbf{Role of the regularization terms.}
We investigate the role of the regularization terms in Eq.~(\ref{eq:finalloss}) by designing several different versions of AD attack and analyzing their attack effectiveness with OOD proxy data.
In what follows, AD-NRA refers to the AD attack without using $ \mathcal{R}(\bm{\theta}_{\mathcal{A}})$ defined in Eq.~(\ref{eq:R_a}),
AD-NLA refers to the AD attack without using $ \mathcal{L}_{a}$ defined in Eq.~(\ref{eq:L_a}), and AD-NRT refers to the AD attack without using any regularization terms, \textit{i.e.}, directly use $\mathcal{L}_\delta+\lambda_{1}\cdot\mathcal{L}_v$ (defined in Eq.~(\ref{eq:pred_lu}) and Eq.~(\ref{eq:loss_v}) respectively) as the objective function.

Figure~\ref{fig:mat_c10} depicts the model performance on the main task after erasing the watermarks by different versions of AD attacks.
From this figure, it is clear that the blue bars, which represent the complete AD, achieve the highest MTA for all considered network architectures.


\section{Conclusion}
This paper has proposed a novel DNN watermark removal attack, AD, at the source data-free regime through continual learning with selective forgetting. 
We introduced two strategies to achieve the attack goal: attention distraction and attention anchoring.
The key to the success of AD is to ensure that the new lure task does not collide with the to-be-forgotten watermark task, and this can be easily achieved by assigning a new label to the lure data. 
Though conceptually simple, extensive experimental results validated that AD outperforms other state-of-the-art works in various benchmark datasets and network architectures. 

\bibliographystyle{IEEEbib}
\bibliography{ref}

\section{Appendix}

\noindent \textbf{Implementation details.}
We evaluate the attack effectiveness of AD on three state-of-the-art watermarks introduced in \cite{zhang2018protecting}, 
including (1) the content watermark (stamp with a self-designed `TEST' pattern on clean samples, (2) the noise watermark (add Gaussian noise on the clean samples), and (3) the unrelated watermark (use the images in MNIST with label `1' as the triggers). 

We set the number of triggers used to embed watermarks into victim models as $1000$, and they are assigned to class $0$, \textit{i.e.}, label `Airplane' in CIFAR-10 and label `Speed Limit 20' in GTSRB. The content watermark triggers and the noise watermark triggers are generated by randomly selecting certain images from each class of the clean training data and stamping the corresponding perturbations on them with the following equation \cite{wang2019neural}:
\begin{equation}
\bm{x}^{(i)}_{t} = (1- \bm{m}) \cdot \bm{x}^{(i)} +  \bm{m} \cdot \bm{\epsilon}^{(i)} 
\label{eq:gen_trigger},
\end{equation}
where $\bm{\epsilon}^{(i)}$ is the trigger pattern determined by $\mathcal{V}$, $\bm{m}$ is a \textit{mask} defines both the position and intensity of the trigger pattern overwritten  into the clean data.

The victim watermarked models with VGG16, ResNet18 and WRN-16-4 are all trained using the Stochastic Gradient Descent (SGD) optimizer with an initial learning rate $0.1$, momentum of $0.9$ and batch size of $128$. And we employed data augmentation techniques, including image rotation, and horizontal and vertical image shift during the training of the victims. 
As for fine-tuning with AD and PST-FT \cite{guo2021hidden}, we use batch size $100$ and no data augmentation strategy are used. The hyper parameters in AD, \textit{e.g.}, 
$\lambda_i~(i=[1, 3])$, are determined by binary search.

\noindent\textbf{Results on GTSRB tasks.}
Table~\ref{tab:gtsrb_results} presents the performance of victim models before and after being attacked by comparing AD with PST-FT on the GTSRB dataset.
As we can see from this table, unlike the results on the CIFAR-10 task, the PST method failed to remove watermarks in most of the victim models on the GTSRB task, \textit{i.e.},  the WMAs in most of the victim models using PST are still higher than the threshold  (since we define the $\epsilon = 0.1$, so for the GTSRB task, the ideal threshold is $12\%$). 
This is mostly because the main task (classification of the GTSRB dataset) and the watermark task of the victim models are easier than those on the CIFAR-10 task, so the victim model is overfitted to the watermark triggers.
For the VGG16 victim with unrelated watermarks, the WMAs almost show no decrease while the MTAs drop about $3\%$ with the PST method. 

Using PST-FT can effectively improve the watermark removal effects. However, most of the victim models using OOD proxy data still cannot make the WMAs below the predefined threshold. 
Using ID proxy data can generally improve the MTAs or reduce the WMAs, and the MTA gaps between the results of using OOD proxy data and ID proxy data are obvious. Similar to what we have analyzed for the experimental results of the CIFAR-10 dataset, it is because of the inherent noise caused by the proxy data distribution mismatch to that of the source domain.

As for AD, it is obvious that all of the three types of watermarks can be removed while the MTAs are close to the victim models before attacks. 
In addition, just like in the CIFAR-10 task, AD attack with ID proxy data performs better than that using OOD proxy data, but the MTA gaps between the results of using OOD proxy data and ID proxy data are small (within $3\%$).

\begin{table*}[]
\caption{Performance of watermark removal attacks on the GTSRB task, where x/y stands for MTA/WMA.}
\label{tab:gtsrb_results}
\begin{tabular}{|c|c|c|c|cc|cc|}
\hline
\multirow{2}{*}{Watermark task} &
  \multirow{2}{*}{Networks} &
  \multirow{2}{*}{Victim models} &
  \multirow{2}{*}{PST} &
  \multicolumn{2}{c|}{PST-FT} &
  \multicolumn{2}{c|}{AD} \\ \cline{5-8} 
 &          &               &               & \multicolumn{1}{c|}{OOD}           & ID            & \multicolumn{1}{c|}{OOD}          & ID           \\ \hline
\multirow{3}{*}{Content} &
  VGG16 &
  96.11 / 100 &
  93.58 / 50.28 &
  \multicolumn{1}{c|}{82.90 / 9.14} &
  91.87 / 0.60 &
  \multicolumn{1}{c|}{95.47 / \textbf{0.00}} &
 \textbf{96.12} / 0.40 \\
 & ResNet18 & 96.30 / 99.48 & 91.55 / 25.08 & \multicolumn{1}{c|}{78.49 / 11.92} & 92.46 / 15.14 & \multicolumn{1}{c|}{92.10 / \textbf{0.04}} & \textbf{95.36} / 8.86 \\
 & WRN-16-4 & 95.66 / 100   & 92.98 / 51.14 & \multicolumn{1}{c|}{81.15 / 11.56} & 92.56 / 26.68 & \multicolumn{1}{c|}{94.97 / 5.08} & \textbf{96.00} / \textbf{2.84} \\ \hline
\multirow{3}{*}{Unrelated} &
  VGG16 &
  96.37 / 100 &
  93.41 / 99.66 &
  \multicolumn{1}{c|}{62.44 / 99.68} &
  92.14 / 0.44 &
  \multicolumn{1}{c|}{94.46 / 0.10} &
  \textbf{96.19} / \textbf{0.00} \\
 & ResNet18 & 96.61 / 100   & 92.29 / 99.98 & \multicolumn{1}{c|}{92.00 / 1.64}  & 92.90 / 2.12  & \multicolumn{1}{c|}{94.98 / \textbf{0.00}} & \textbf{96.17} / \textbf{0.00} \\
 & WRN-16-4 & 95.28 / 100   & 91.28 / 99.80 & \multicolumn{1}{c|}{80.18 / 99.94} & 87.73 / 94.02 & \multicolumn{1}{c|}{\textbf{94.84} / \textbf{0.34}} & 94.74 / 2.38 \\ \hline
\multirow{3}{*}{Noise} &
  VGG16 &
  96.34 / 100 &
  93.72 / 10.06 &
  \multicolumn{1}{c|}{93.31 / 3.80} &
  93.00 / 3.68 &
  \multicolumn{1}{c|}{95.69 / 0.30} &
  \textbf{96.30} / \textbf{0.00} \\
 & ResNet18 & 95.44 / 99.98 & 92.90 / 0.72  & \multicolumn{1}{c|}{92.38 / 0.26}  & 92.52 / \textbf{0.00}  & \multicolumn{1}{c|}{94.10 / \textbf{0.00}} & \textbf{95.14} / \textbf{0.00} \\
 & WRN-16-4 & 95.79 / 100   & 92.11 / 47.34 & \multicolumn{1}{c|}{82.31 / 11.40} & 82.60 / 2.30  & \multicolumn{1}{c|}{93.00 / \textbf{0.00}} & \textbf{95.88} / 3.48 \\ \hline
\end{tabular}
\end{table*}

\begin{table*}[h]
\caption{Performance of watermark removal attacks of AD without using lure data on GTSRB.}
\label{tab:vanilla_aa1}
\centering
\begin{tabular}{|c|cc|cc|}
\hline
\multirow{2}{*}{Networks} & \multicolumn{2}{c|}{Vanilla}                       & \multicolumn{2}{c|}{AA}                           \\ \cline{2-5} 
                          & \multicolumn{1}{c|}{OOD}           & ID            & \multicolumn{1}{c|}{OOD}           & ID           \\ \hline
VGG16                     & \multicolumn{1}{c|}{58.07 / 19.78} & 94.39 / 12.04 & \multicolumn{1}{c|}{57.48 / \textbf{1.72}}  & \textbf{95.67} / 3.80 \\
ResNet18                  & \multicolumn{1}{c|}{66.29 / 50.48} & 95.19 / \textbf{0.16}  & \multicolumn{1}{c|}{89.23 / 21.68} & \textbf{95.41} / 0.20 \\
WRN-16-4                  & \multicolumn{1}{c|}{85.37 / 99.98} & \textbf{95.37} / 8.20  & \multicolumn{1}{c|}{79.13 / 99.90} & 95.31 / \textbf{1.90}\\ \hline
\end{tabular}
\end{table*}

\noindent \textbf{Role of the lure task.}
Table~\ref{tab:vanilla_aa1} list the results using vanilla attack and AA attack on various victim models with content-based watermark.
We can observe from the table that, when using OOD proxy data, neither the vanilla attack nor the AA attack is as good as AD in solving the SFW.
Specifically, for VGG16, although the WMA of victim models drops to below $12$\%, their MTAs also drop drastically after attacks (as low as $58.07$\% for vanilla attack and $57.48$\% for AA attack). 
And in both attacks, the gap in WMA for OOD and ID proxy data is huge (with the largest gap being about $38$\%), while this value is only $3$\% for AD as shown in Table~\ref{tab:gtsrb_results}. 
This, once again, corroborates the lure task plays a key role in solving SFW: the lure task will compensate for the negative impact caused by the distribution difference between the proxy data and the source training data. 

\begin{figure*}[b]
  \centering
  \includegraphics[width=\linewidth]{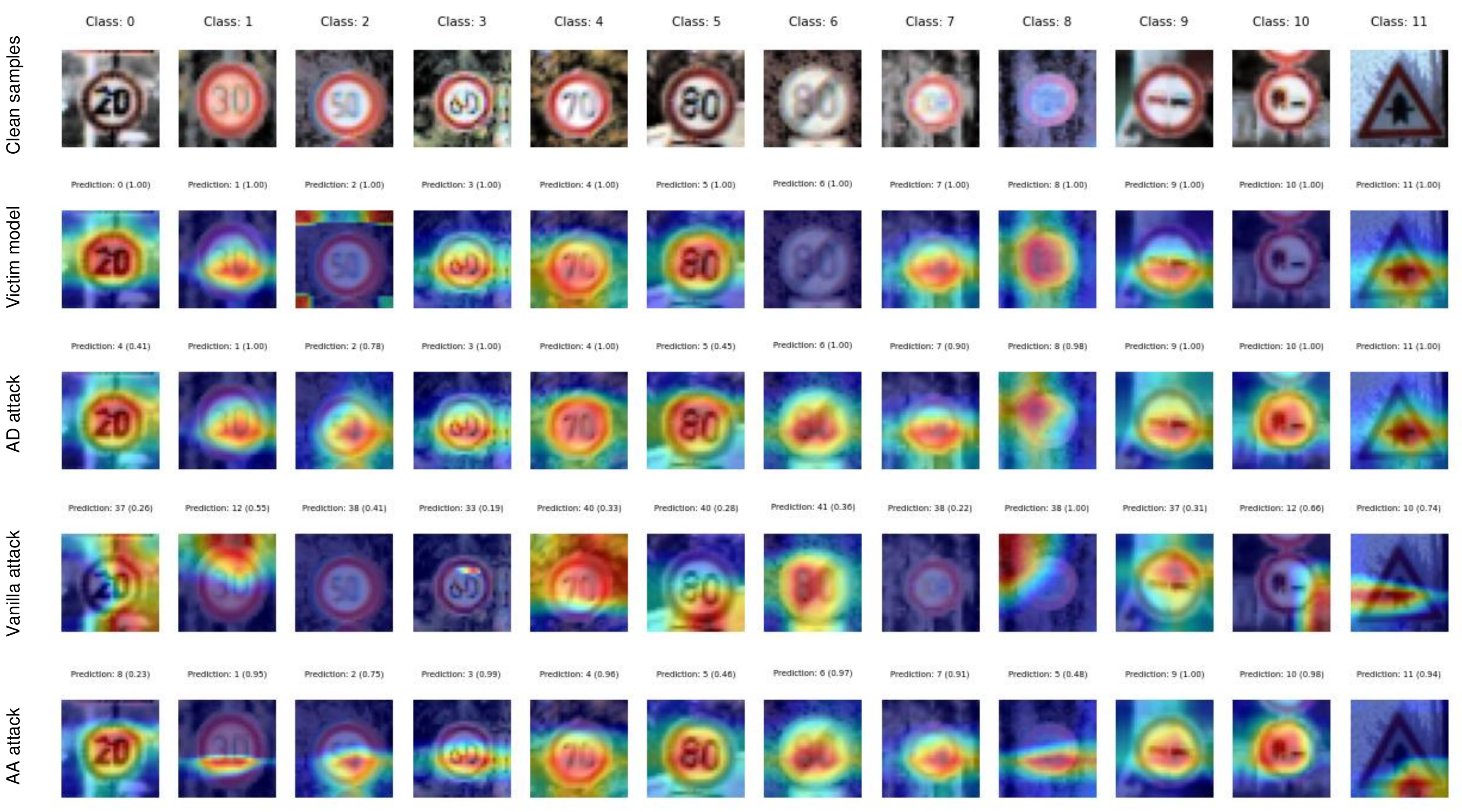}
  \caption{The attention heatmaps of ResNet18 on part of GTSRB clean instances. The first row depicts the clean images from different classes. 
The second row depicts the heatmaps of the victim model on the clean images. The following rows respectively depict the heatmaps of the model after the AD attack, vanilla attack, and AA attack with OOD proxy data.}
\label{fig:heatmaps}
\end{figure*}

The heatmaps of the victim models before and after attacks on clean instances can further explain the role of the lure task in a more vivid form.
As presented in Fig.~\ref{fig:heatmaps}, which visualizes the attention heatmaps of ResNet18, before and after different attacks using OOD proxy data, on GTSRB images.
From the second and third rows of this figure, it is easy to see that the attention maps of the model after the AD attack are highly similar to those of the victim before the attacks. 
For the vanilla and AA attacks, the attention maps of the model after attack deviate from those of the victim before attack. 
That said, without the participation of lure, using OOD proxy data alone cannot effectively anchor the main task attention maps while forgetting the target watermark task. 

\begin{table}[]
\caption{Influence of the lure data label on GTSRB.}
\label{tab:label}
\begin{tabular}{|c|c|c|c|}
\hline
Network  & {$\Delta=0$}      &{$\Delta=10$}     & {$\Delta=20$}             \\ \hline
VGG16    & 69.00 / 99.96 & \textbf{95.64} / 5.12  & 95.47 / \textbf{0.64} \\
ResNet18 & 79.88 / 96.50 & \textbf{92.24} / 7.98  & 92.04 / \textbf{0.88} \\
WRN-16-4 & 61.18 / 100   & 93.76 / 11.18 & \textbf{94.31} / \textbf{3.10} \\ \hline
\end{tabular}
\end{table}

\begin{figure}[t]
\centering
\subfigure[Content]{
\includegraphics[width=3.9cm]{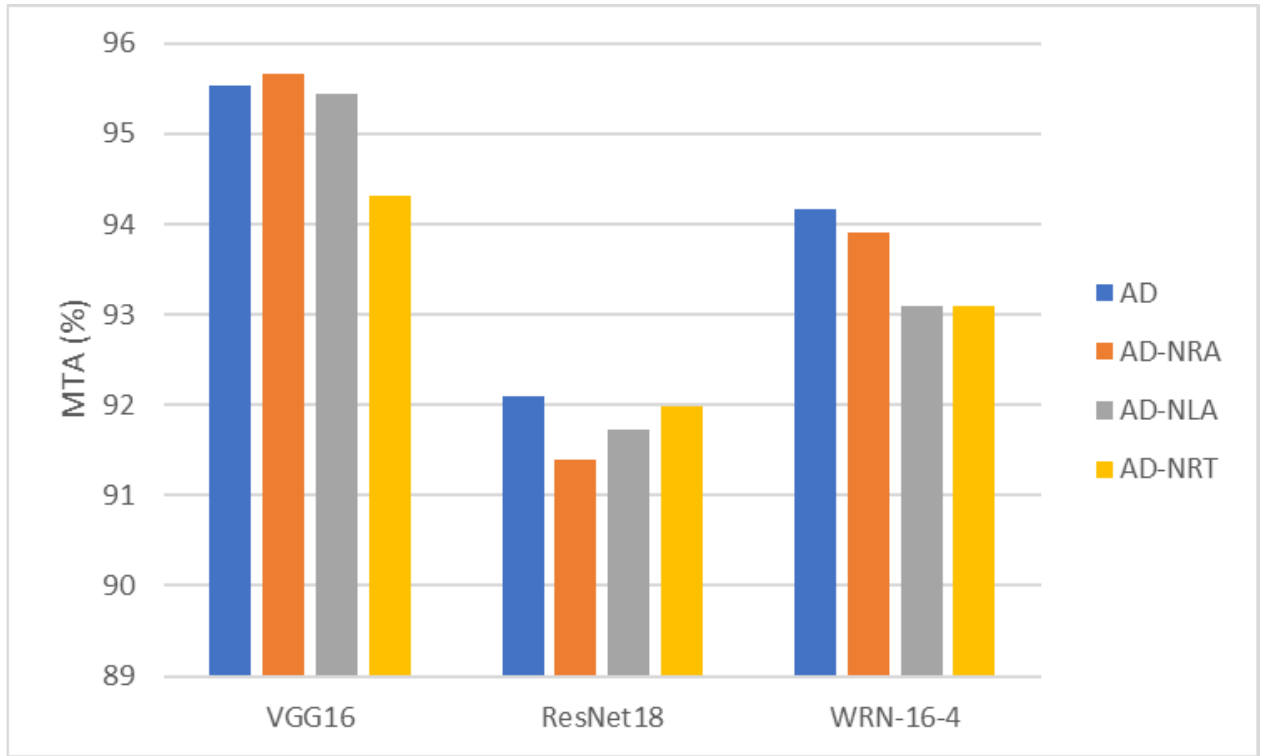}
}
\subfigure[Unrelated]{
\includegraphics[width=3.9cm]{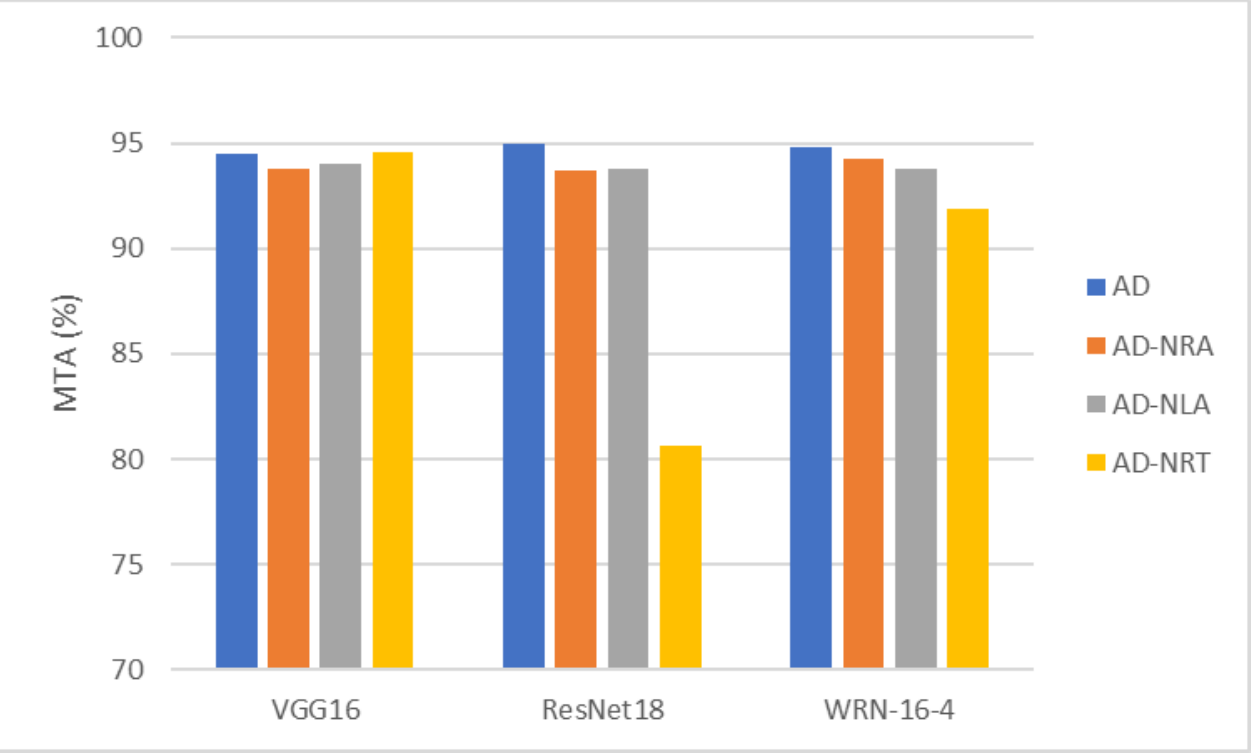}
}
\caption{The MTAs of regularization terms on the GTSRB task, where (a) refers to the content-based watermark, (b) refers to the unrelated-based watermark.
Their corresponding WMA is lower than the predefined threshold. The blue bars represent the complete version of AD, the orange bars, grey bars, and yellow bars represent different versions of AD, respectively.}
\label{fig:RT_gtsrb}
\end{figure}

\noindent\textbf{Influence of the label of the lure task.}
The experiment results using different predefined lure labels are listed in Table~\ref{tab:label}.
We can observe from this table that when the lure label $\Delta$ is not overlap with the target label $0$, \textit{i.e.}, $\Delta=10$ or $20$, the results of AD attack is comparable to that  using a new class $C$ as the label for lure data. 
Specifically, when $\Delta$ is $10$ or $20$,  the WMA of the AD-attacked models is below $12$\% and the loss in MTA is small.
However, when $\Delta=0$, \textit{i.e.,} the label of the lure task overlaps with the target label, the AD attack fails for all victim models.
This phenomenon once again verifies the necessity of using an additional defined new label for the lure task when the watermark is agnostic, \textit{i.e.}, preventing label collision.

\begin{figure}[b]
\centering
\subfigure[Content]{
\includegraphics[width=2.5cm]{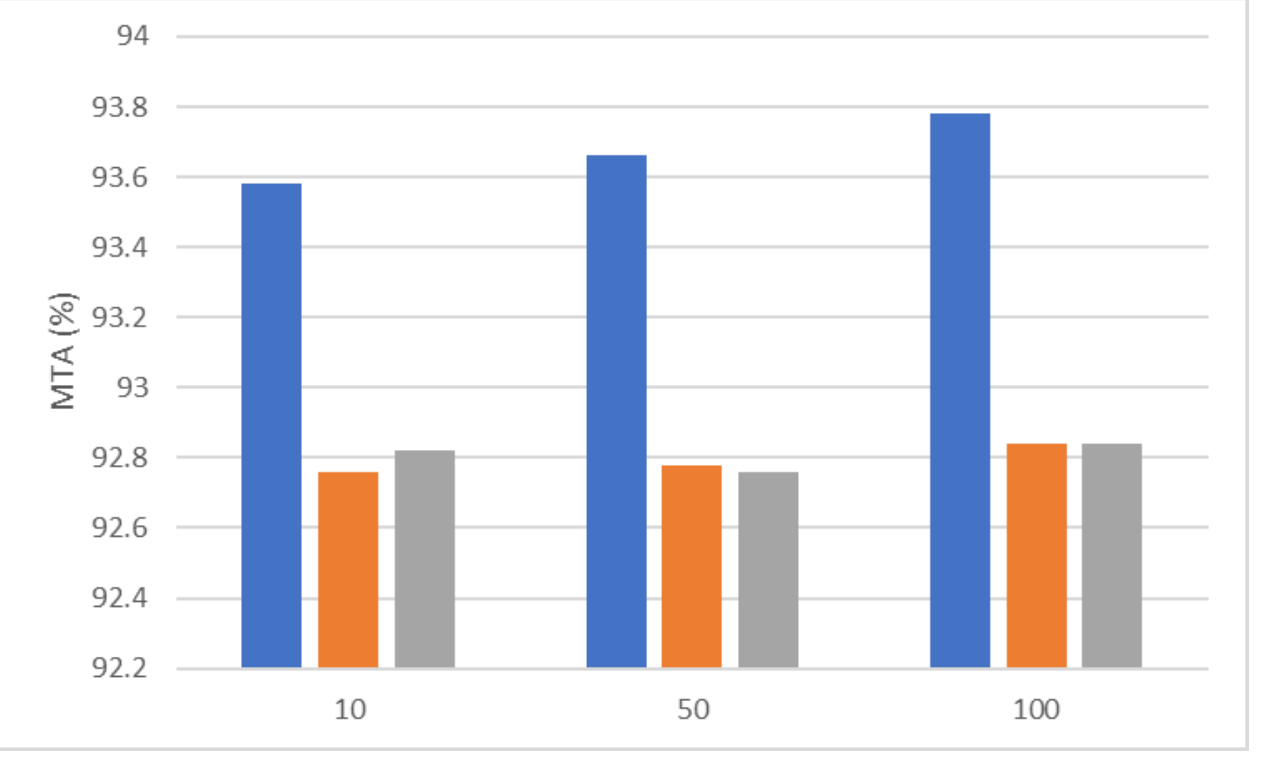}
}
\subfigure[Unrelated]{
\includegraphics[width=2.5cm]{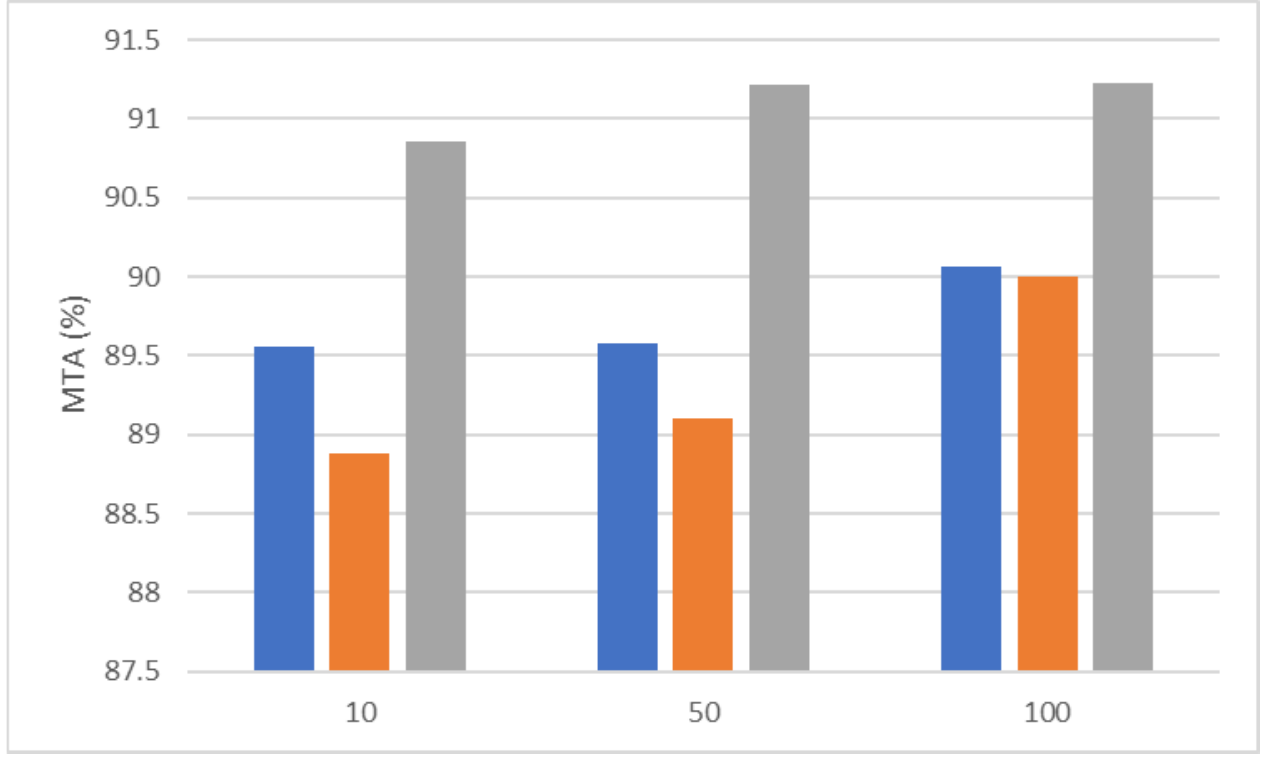}
}
\subfigure[Noise]{
\includegraphics[width=2.5cm]{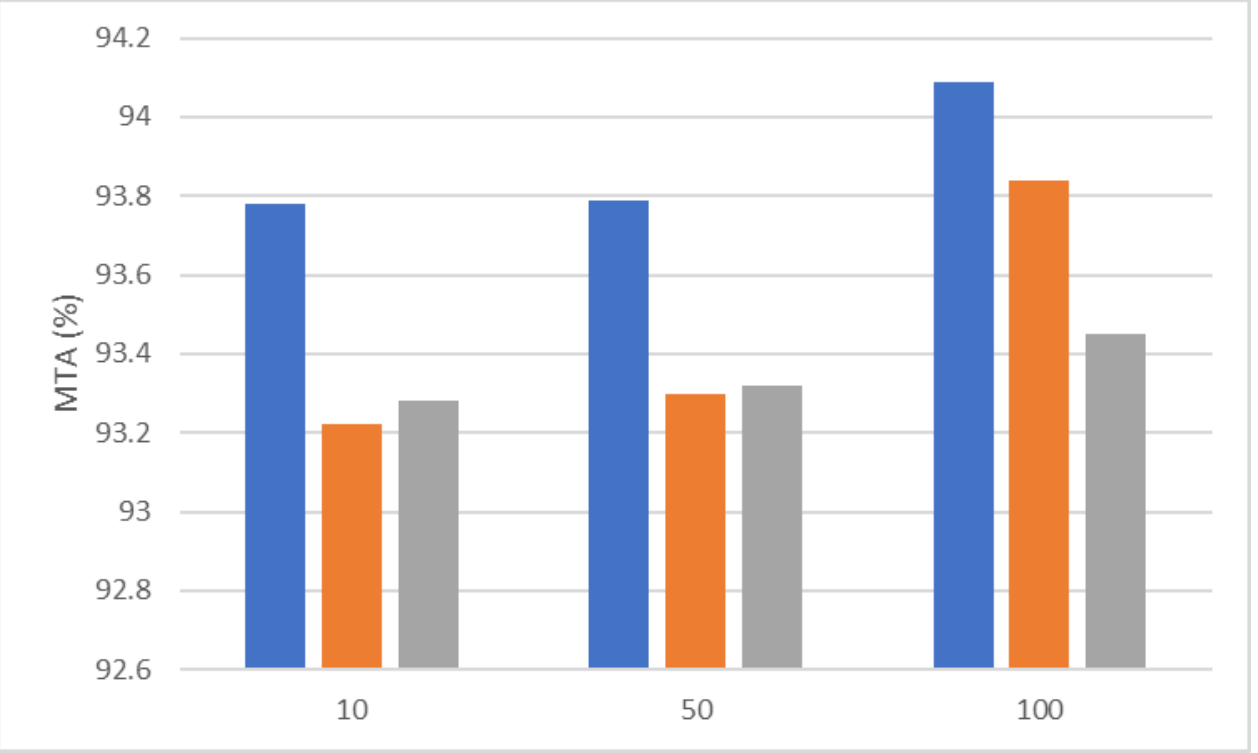}
}
\quad
\subfigure[Content]{
\includegraphics[width=2.5cm]{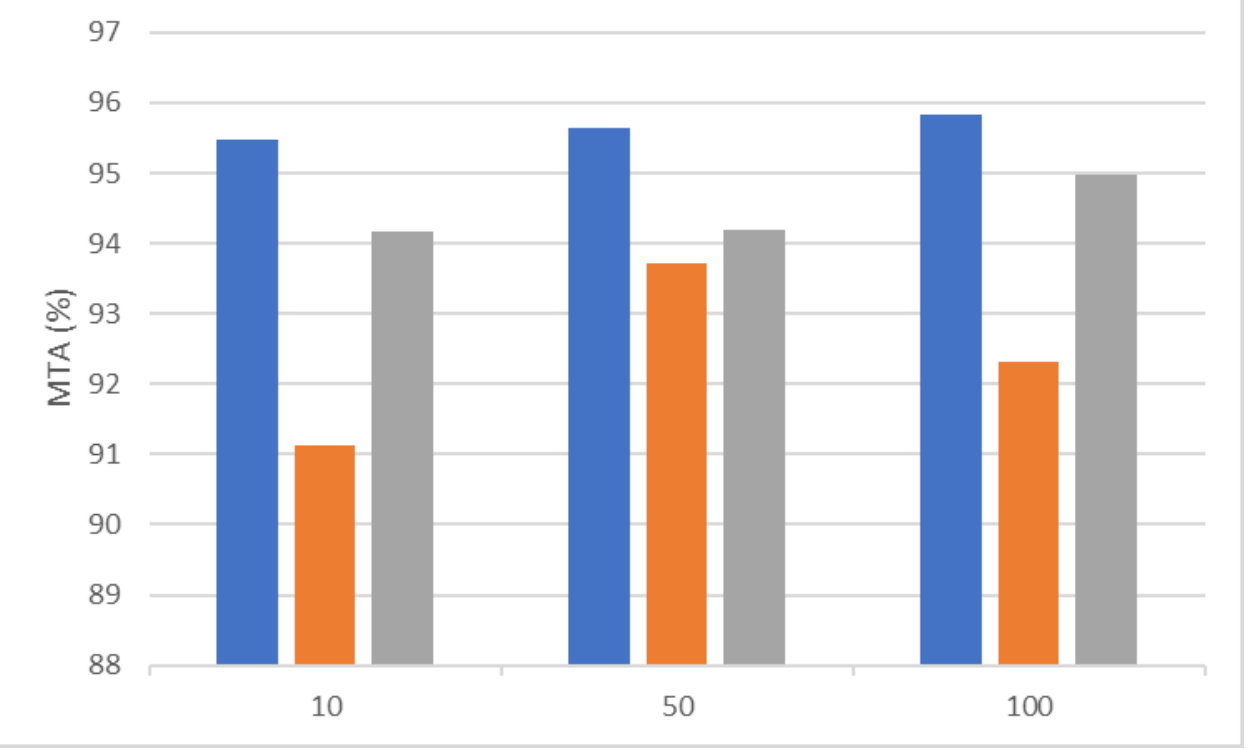}
}
\subfigure[Unrelated]{
\includegraphics[width=2.5cm]{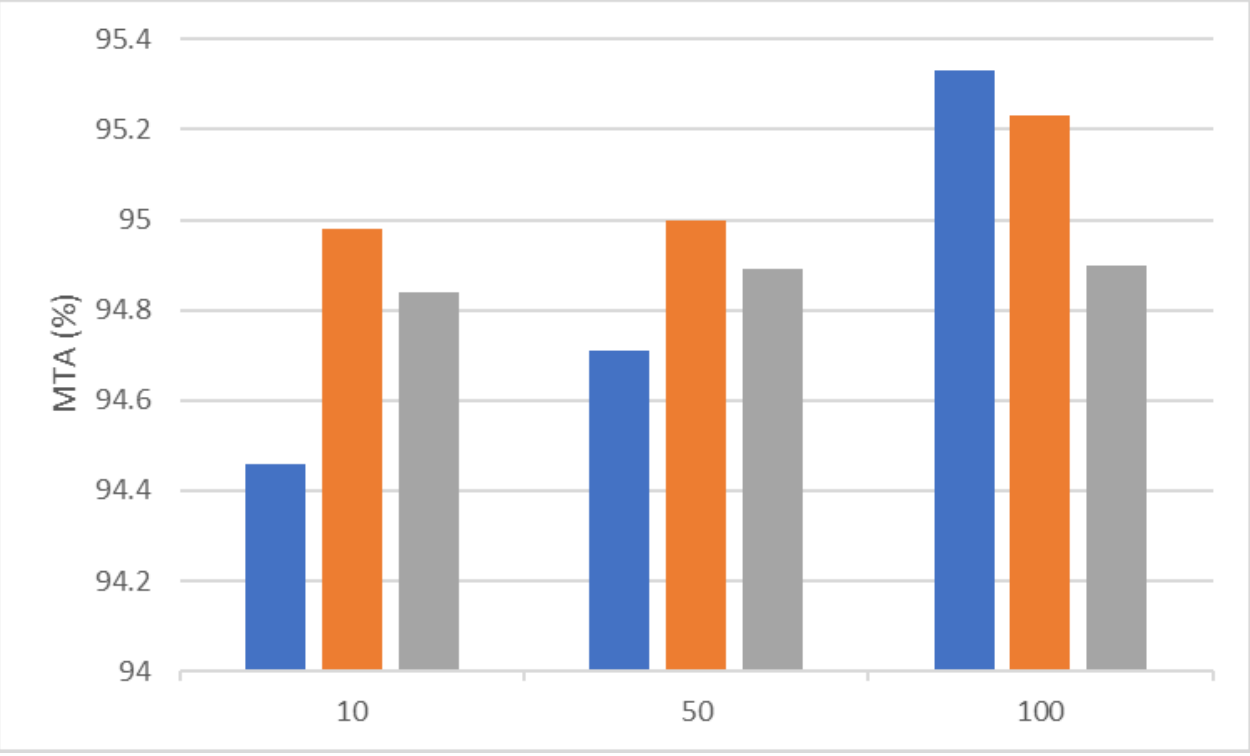}
}
\subfigure[Noise]{
\includegraphics[width=2.5cm]{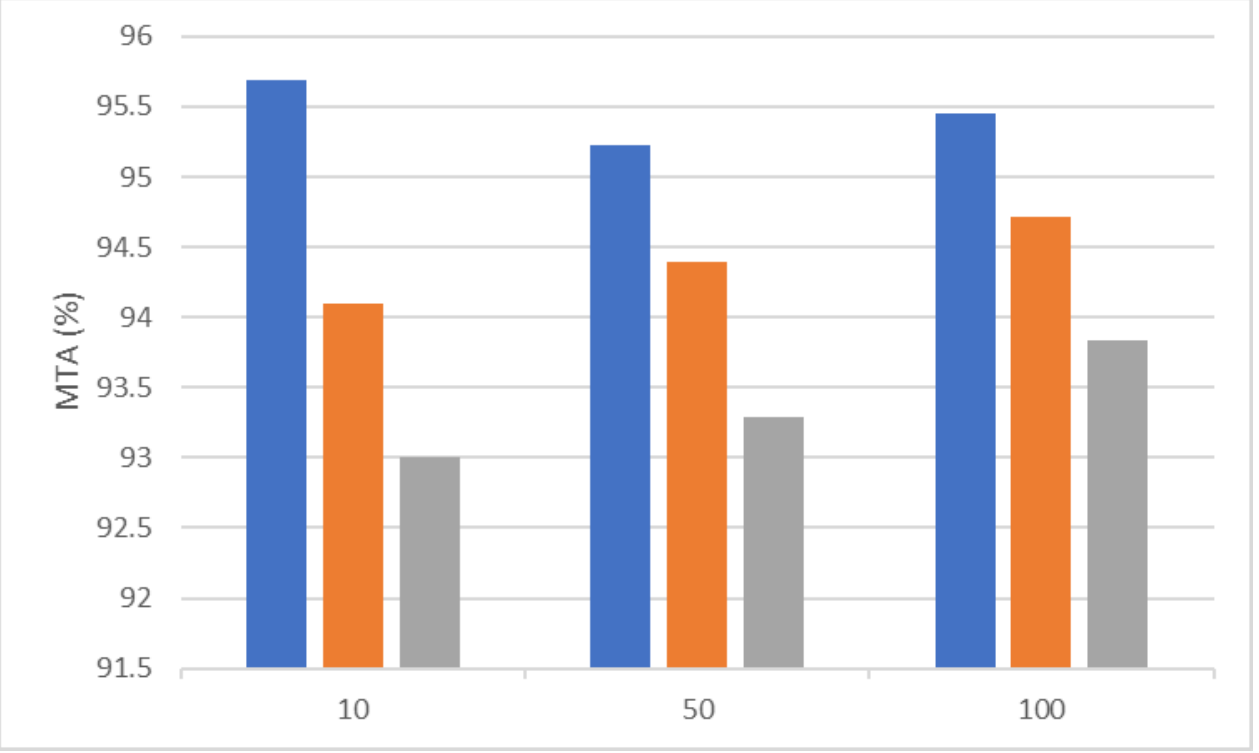}
}

\caption{Sensitivity to the lure data budget on different victim models, where the lure data used is abstract images, (a)-(c) refer to the models on CIFAR-10 task, (d)-(f) refer to the models on GTSRB task. The blue bars, orange bars, and grey bars represent the VGG16, ResNet18, and WRN-16-4 networks, respectively. Their corresponding WMAs are below the threshold.}
\label{fig:lure_abs}
\end{figure}

\noindent\textbf{Influence of lure data budget.}
We also investigate the attack effects of different lure data budgets (using abstract images as the lure data), and the results are depicted in Fig.~\ref{fig:lure_abs}.
We can see that there is an improvement in the MTA of most victim models as the amount of  lure data increases, even though the improvement is limited (within $1\%$).
This indicates that the proposed AD is a serious threat to existing DNN watermarking methods.

\begin{figure}[h]
\centering
\subfigure[VGG16]{
\includegraphics[width=2.5cm]{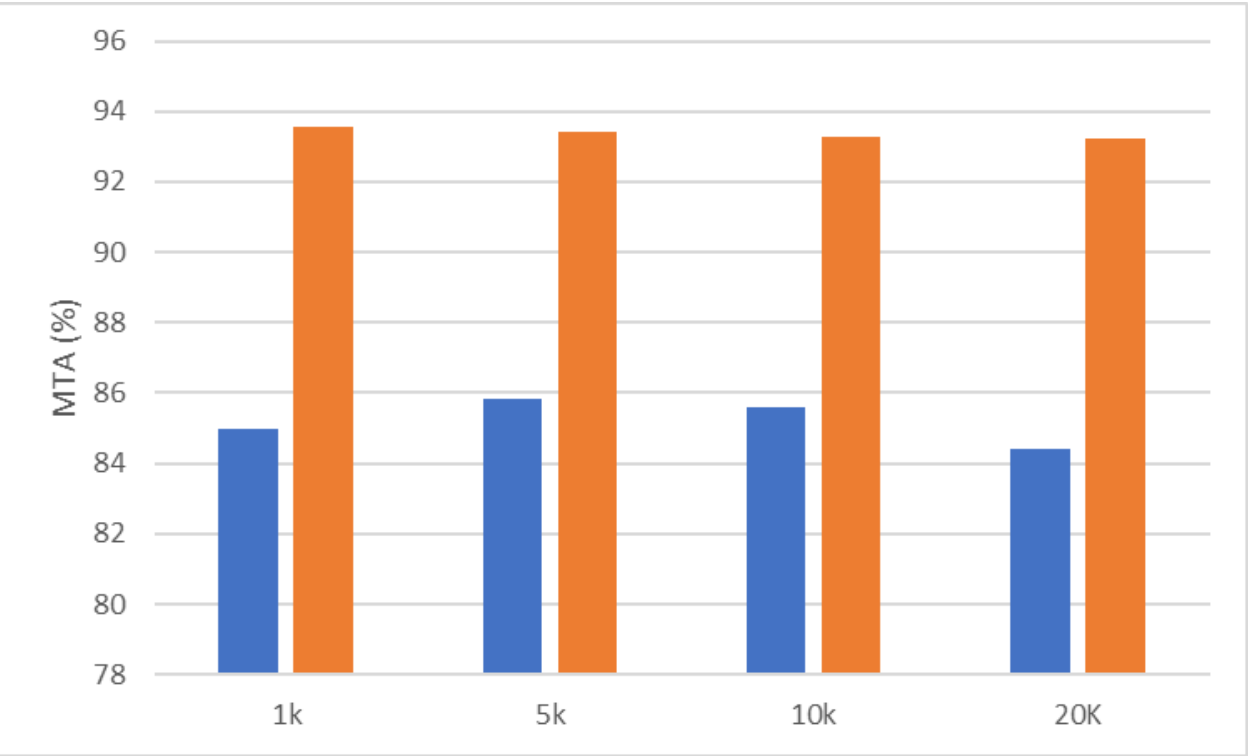}
}
\subfigure[ResNet18]{
\includegraphics[width=2.5cm]{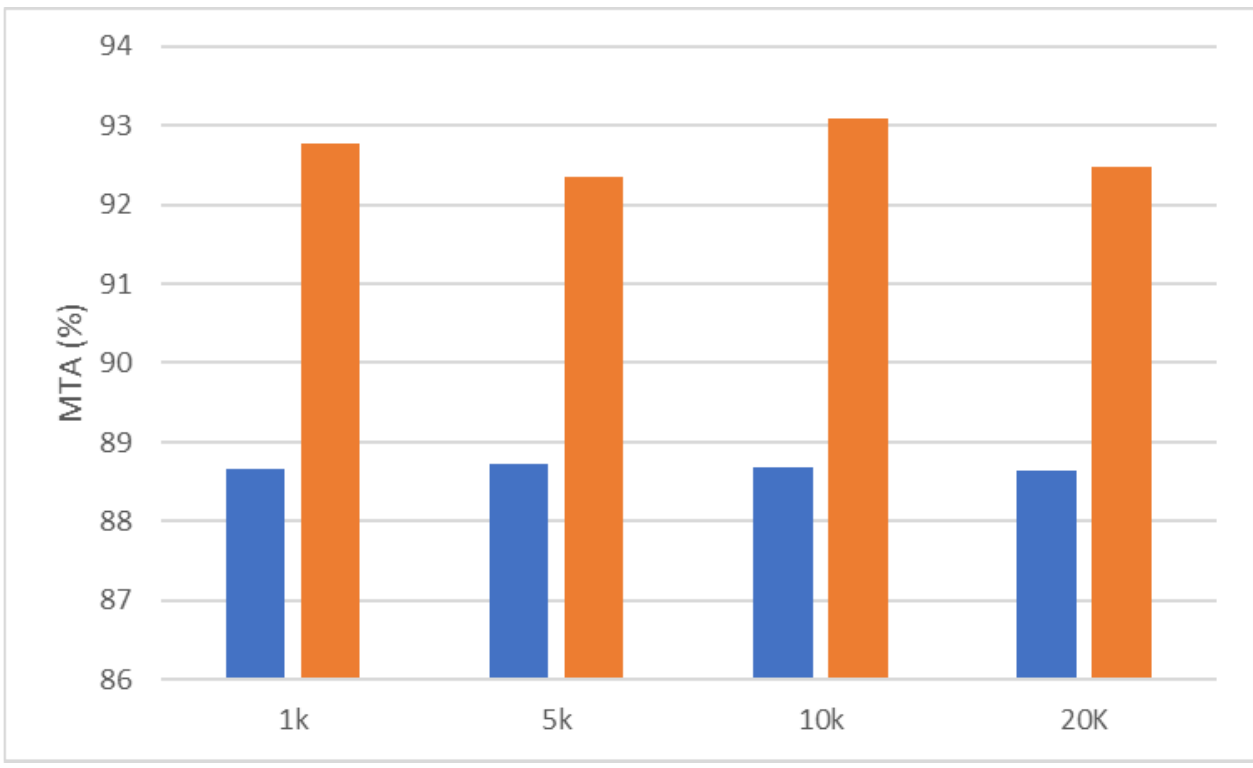}
}
\subfigure[WRN-16-4 ]{
\includegraphics[width=2.5cm]{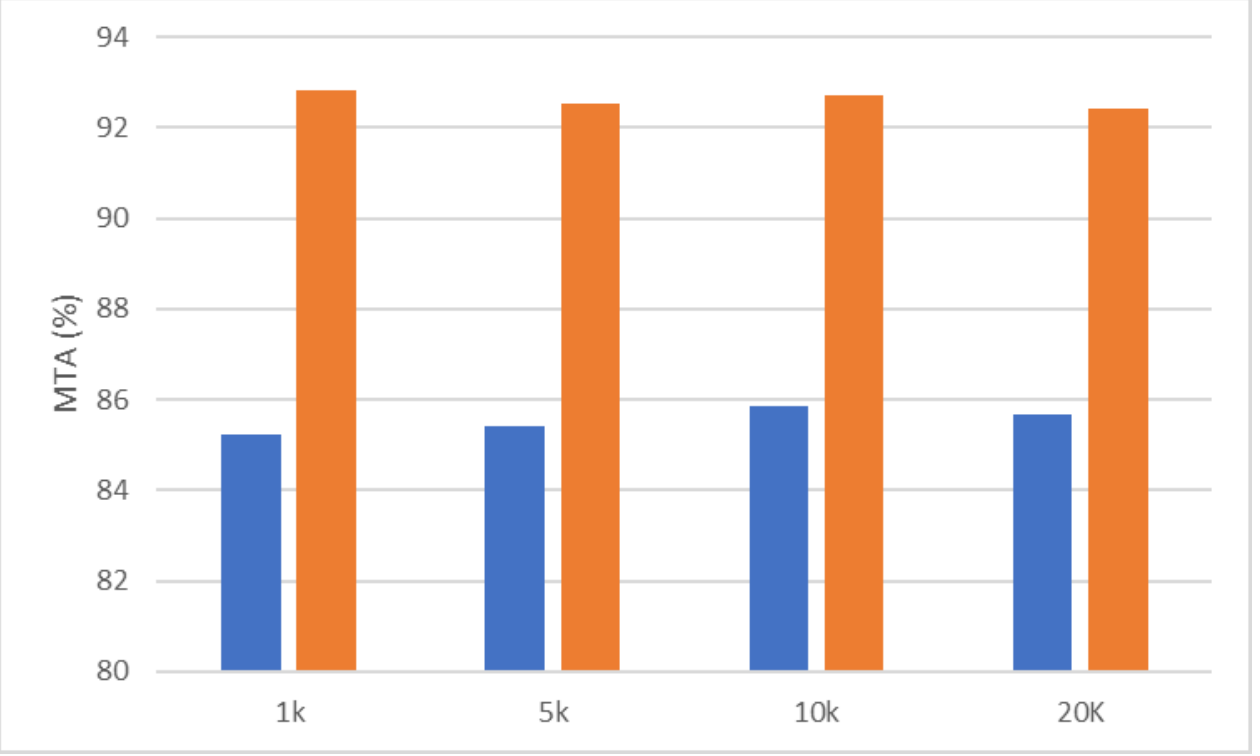}
}
\quad
\subfigure[VGG16]{
\includegraphics[width=2.5cm]{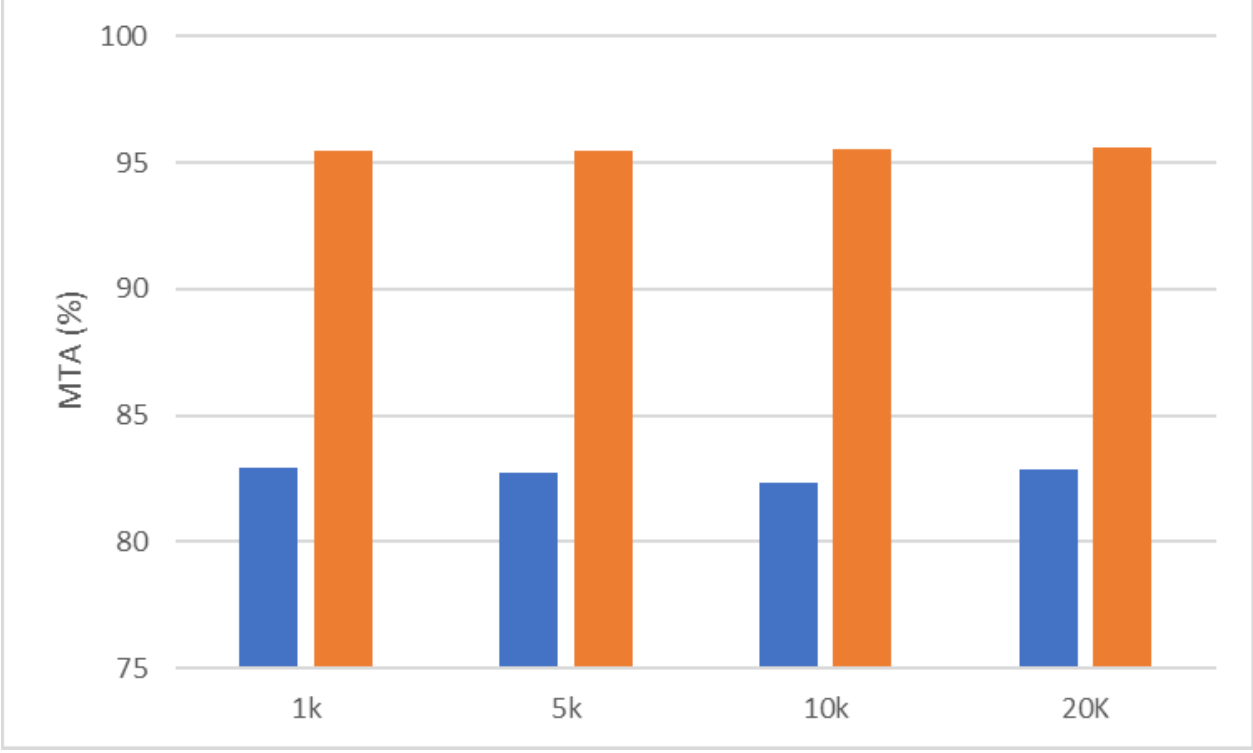}
}
\subfigure[ResNet18]{
\includegraphics[width=2.5cm]{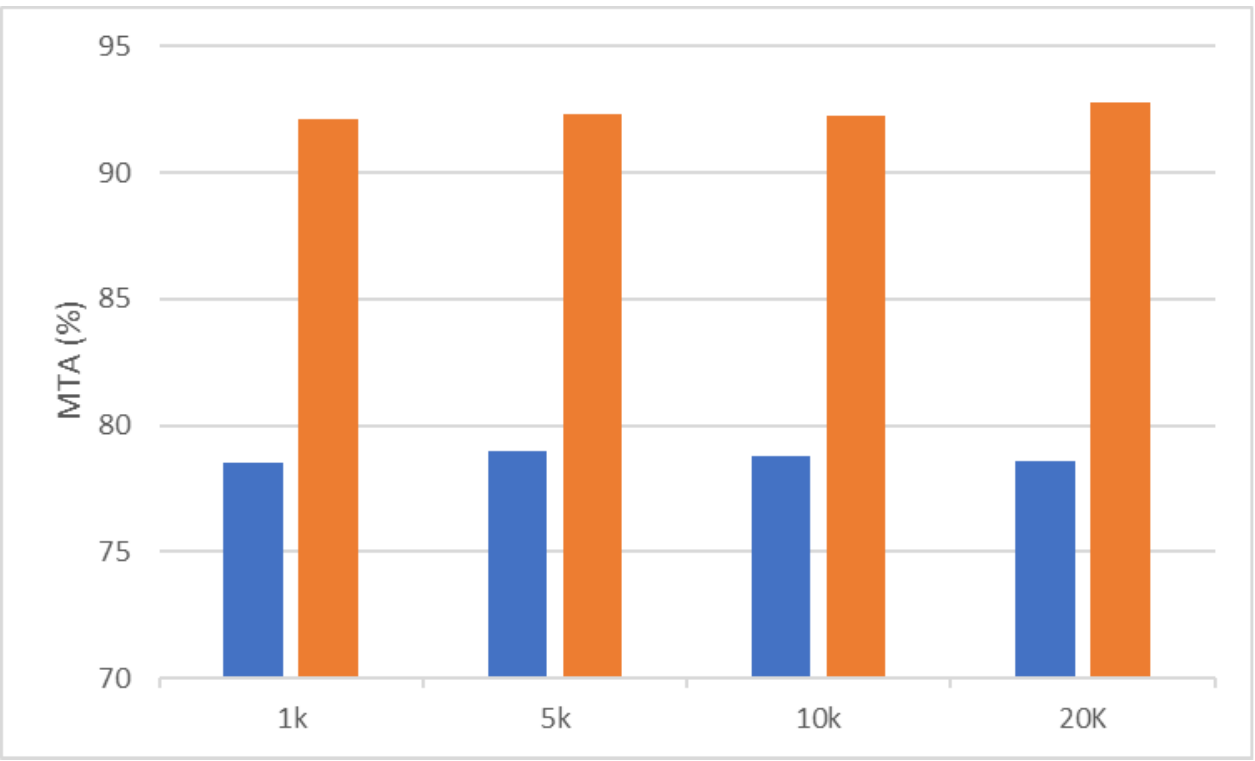}
}
\subfigure[WRN-16-4 ]{
\includegraphics[width=2.5cm]{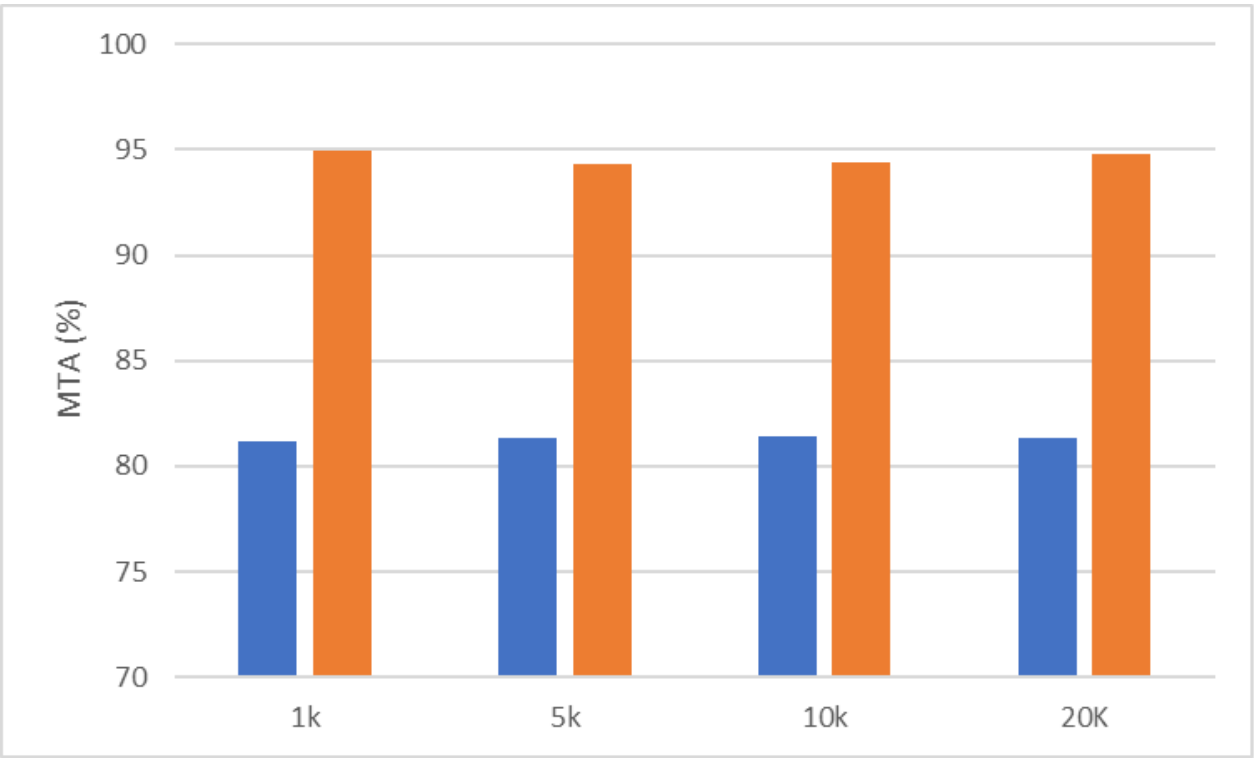}
}

\caption{Performance of AD attack and PST-FT attack with different amount of OOD proxy data, where (a)-(c) refer to the models on CIFAR-10 task, (d)-(f) refer to the models on GTSRB task. The blue bars represent the PST-FT attack, and the orange bars represent the AD attack. Their corresponding WMAs are below the corresponding thresholds.}
\label{fig:proxy}
\end{figure}

\noindent\textbf{Influence of proxy data budget.}
We evaluate the performance of AD and PST-FT with different proxy data budgets on content-based watermarks, as shown in Fig.~\ref{fig:proxy}.
We observe that for both the AD attack and the PST-FT attack, the MTAs of the victim models are preserved with the increase of the volume of the proxy data. And for some networks, their MTAs have slightly increased when using more proxy data in fine-tuning. However, the MTA gaps between PST-FT and AD are still obvious even using more proxy data.

\noindent\textbf{Role of the regularization terms.}
Figure~\ref{fig:RT_gtsrb} depicts the model performance on the main task (GTSRB) after different versions of AD attacks.
From the above figure, we can see that almost all the blue bars, which represent the AD, reach the highest MTA.
For the VGG16, the attack effects of AD-NPA and AD-NAA are almost the same, reaching the complete version of AD. 
By comparing the orange bars, grey bars, and yellow bars, for most of the networks and watermarks, AD-NPA and AD-NAA are better than AD-NRT.
For the ResNet18 and WRN-16-4, the difference in attack effects between different versions of AD is more obvious.
Specifically, for the ResNet18 with the unrelated-based watermark, the attack effects of AD-NRT are obviously less than those of the other versions of AD.
By comparing the blue and yellow bars, we find that the attention loss term and the attention penalty term help to maintain the main task performance. 
Employing the prediction alignment is not sufficient to achieve good performance in MTA.

\end{document}